\documentclass[12pt]{article}

\usepackage{preprint-layout}
\usepackage{preprint-notation}

\usepackage{subcaption}
\usepackage{hyperref}
\usepackage{import}
\usepackage{multirow}
\usepackage{amssymb}
\usepackage{siunitx}

\DeclareMathOperator*{\argmin}{\arg\min}
\ppnewcommand{\bfxi}{\boldsymbol \xi}

\title{Learning Nonlinear Finite Element Solution Operators using Multilayer Perceptrons and Energy Minimization}
\date{\today}

\author{
Mats G. Larson,
Carl Lundholm,
Anna Persson
}
\date{}

\begin{document}

\maketitle

\begin{abstract}
We develop and evaluate a method for learning solution operators to nonlinear problems governed by partial differential equations (PDEs). The approach is based on a finite element discretization and aims at representing the solution operator by a multilayer perceptron (MLP) that takes problem data variables as input and gives a prediction of the finite element solution as output. The variables will
typically correspond to parameters in a parametrization of input data such as boundary conditions, coefficients, and right-hand sides. The output will be an approximation of the corresponding finite element solution, thus enabling support and enhancement by the standard finite element method (FEM) both theoretically and practically.
We focus on using the corresponding energy functional to a PDE as the loss function since this allows us formulate
efficient parallelizable training algorithms based on assembling the energy locally on each element. For large problems, the learning process can thus be made more efficient by using only a small fraction of randomly chosen elements in the mesh in each iteration.
The method is evaluated on several relevant test cases, where learning the finite element solution operator turns out to be beneficial, both in its own right but also by combination with standard FEM theory and software.
\end{abstract}

\section{Introduction}
In recent years, there has been a lot of interest in using machine learning methods in applied mathematics and scientific computing, see for example \cite{bookDLnCP}. In this work, we focus on physics-informed methods that approximate the solution operator to parameterized partial differential equations (PDEs) that admit an energy functional. We emphasize that the aim is \textit{not} to learn the solution to a single PDE problem, but rather to a family of the same type of PDE problems, by learning the mapping from a low-dimensional parameter space into the solution space. As the solution space is typically infinite-dimensional we consider \emph{discrete} spaces associated with standard numerical methods for PDEs. We thus aim at learning well-known approximations instead of the exact solution. The idea is that the standard numerical method could support and enhance the machine learning one. Here we consider the finite element method (FEM) and present some examples of beneficial combination with the machine learning framework.

In the following literature review we start off broadly and then successively narrow down the works to ones with more and more similarities to our own approach. After a general introduction, we focus on physics-informed methods, then move on to approaches for parameterized PDEs, followed by FEM-NN hybrids where we focus on graph-based methods, finally we make some direct comparisons between our method and the most similar ones found in the literature. We point out that the field of learning solutions to PDEs is growing fast so the works mentioned here should not be considered an extensive list and that relevant contributions might have been overlooked.

There are several approaches to learning solution operators or solutions to PDEs in the literature, see for example \cite{MR4582511} where neural operators (lifting map, followed by a modified MLP in the sense that a kernel integral operator is added to the weights-bias sum, and ending with a projection map) on a general integral form together with some concrete examples are presented. The different approaches can be divided into two main branches; data-driven or physics-informed. The data-driven methods are based on (typically pre-computed) data, such as, e.g., the neural operators in \cite{Stuart_NO}, Fourier neural operators \cite{Stuart_FNO}, DeepONets \cite{Lu_et_al_DeepONets}, and the random feature model \cite{Nelsen_Stuart_21}. The physics-informed methods typically penalize the network to satisfy a PDE or other physical laws by designing an appropriate loss function, e.g., an energy-based loss function such as in the Deep-Ritz method \cite{E_etal_2018} or a loss function based on the strong residual as for physics-informed neural networks (PINNs) \cite{Raissi_etal_2019}. These methods generally do not require any data in the classical machine learning sense. It is worth noting that the earliest examples of physics-informed neural methods for differential equations were presented in the 1990s with works such as \cite{LEE1990110}, where minimization of finite difference residuals is considered, and \cite{DissanayakePhanThien1994, Lagaris1997ArtificialNN} which are closer to modern-day PINNs.

Typically, the focus of physics-informed approaches has not been on learning parameterized problems, but mainly the solution to a single PDE problem, often with collocation type methods such as \cite{E_etal_2018, Raissi_etal_2019, brink_neural_2021, mishra_artificial_2024, weng_deep_2025}. Recently however, works on learning parameterized PDEs seem to have gained more traction since they allow for a single network to predict the solutions to a class of related problems, see, e.g., \cite{URIARTE2022114562, MR4645137, heiss2023, khara_neufenet_2024}, or \cite{MR4597397, cho2024} where PINNs is extended to parameterized settings, and \cite{MR4467422} in which networks are trained to parameterize the Green function. Several parameterized-PDE approaches also take into account the variational formulation in some manner: \cite{bachmayr2024} in a loss function related to the error in the norm induced by the variational formulation, \cite{MR4756922} by basing the the loss function on the weak residual, and \cite{PATEL2024116536} instead in the network architecture.

Combining FEM with neural networks is not new. See for example \cite{Guan2023, Bergel_2023, ABALLAY2026114447} or \cite{MITUSCH2021110651} for hybrid FEM-NN models where networks are used to model unknown components of the PDE, e.g., terms and coefficients. Another example is \cite{THEL2024117073} in which networks are used to replace regions of the finite element mesh to speed up crash simulations. A key feature of FEM is the flexibility in handling complex irregular geometries by allowing for unstructured meshes. Using the natural connection between graphs and unstructured meshes, graph-based FEM learning approaches have been proposed. An influential contribution is the physics-informed graph Galerkin neural networks \cite{GAO2022114502} that predict finite element solutions to stationary parameterized PDEs by using graph convolutional networks (GCNs) and weak residual minimization. A FEM-residual loss is also considered in \cite{WURTH2024117102} to solve time-dependent problems with MeshGraphNets, a combination of an MLP, a message passing network, and a convolutional neural network (CNN). For methods focusing on elasticity, see \cite{He2023} which combines GCNs with energy minimization, and \cite{DESHPANDE2024108055} where a graph-based encoder-decoder network is used.

The most similar works to our method that we have found in the literature are \cite{Meethal2023, MR4645137, khara_neufenet_2024}:

\begin{itemize}
\item The FEM-NN hybrid presented in \cite{Meethal2023} also maps parameter tuples to the finite element degrees of freedom (DOFs), but the loss function is based on the weak residual, not energy as here, and only linear and relatively small problems are considered.

\item In \cite{MR4645137} mesh-informed neural networks (MINNs) are used to again map parameter tuples into a finite element space. There the loss function is instead based on the strong residual and there is no analysis, but several impressive examples are presented.

\item Also in \cite{khara_neufenet_2024} the map from a parameter space into a finite element space is learned, and some analysis is presented together with examples. However, there a field-to-field map is considered with CNNs (a natural choice for such a map) and both the analysis and examples are for linear problems. In our method we consider a tuple-to-field map with MLPs and both analysis and examples include nonlinearities (the tuple contains the parameters, which do not need to be associated with points in the solution domain, as opposed to for a field).
\end{itemize}

In this paper we consider a data-free physics-informed method for operator learning of parameterized PDEs. The method is actually a special case of the more general approach presented in \cite{sharp2023} and it has already been applied to inverse problems in \cite{BURMAN2025118111}. However, in the former, the method is barely noticeable in the more general framework presented there. In the latter, the focus lies on the application rather than on the method. Here, we therefore put full attention on the method itself, develop and present it in a general form along with some analysis and examples.

We construct a neural network that learns a family of solutions by utilizing a loss function based on energy minimization that penalizes the network to satisfy the PDE. This means that the computationally costly step of generating training data is avoided. The energy minimization approach has earlier been considered in neural network contexts, e.g., \cite{E_etal_2018, samaniego_energy_2020, Li2021, ESHAGHI2025} and \cite{sharp2023} for finding low-dimensional models of physical motion and \cite{Sacks2022} for applications in soft tissue modeling. In this work we consider a method that can be applied to any (nonlinear) problem with a given energy functional. The problem is first discretized using a finite element method and a multilayer perceptron (MLP) neural network is trained to learn the corresponding approximate solution operator. The input to the neural network is parameters to the problem that define PDE problem data, e.g., boundary conditions, coefficients, or right-hand side. The output of the network is the finite element degrees of freedom, which for, e.g., classical P1 elements are the function values at the nodes of the finite element mesh.

\begin{figure}
\centering
\def\svgwidth{0.49\textwidth}
\begingroup%
  \makeatletter%
  \providecommand\color[2][]{%
    \errmessage{(Inkscape) Color is used for the text in Inkscape, but the package 'color.sty' is not loaded}%
    \renewcommand\color[2][]{}%
  }%
  \providecommand\transparent[1]{%
    \errmessage{(Inkscape) Transparency is used (non-zero) for the text in Inkscape, but the package 'transparent.sty' is not loaded}%
    \renewcommand\transparent[1]{}%
  }%
  \providecommand\rotatebox[2]{#2}%
  \newcommand*\fsize{\dimexpr\f@size pt\relax}%
  \newcommand*\lineheight[1]{\fontsize{\fsize}{#1\fsize}\selectfont}%
  \ifx\svgwidth\undefined%
    \setlength{\unitlength}{473.76901567bp}%
    \ifx\svgscale\undefined%
      \relax%
    \else%
      \setlength{\unitlength}{\unitlength * \real{\svgscale}}%
    \fi%
  \else%
    \setlength{\unitlength}{\svgwidth}%
  \fi%
  \global\let\svgwidth\undefined%
  \global\let\svgscale\undefined%
  \makeatother%
  \begin{picture}(1,0.55855055)%
    \lineheight{1}%
    \setlength\tabcolsep{0pt}%
    \put(0,0){\includegraphics[width=\unitlength,page=1]{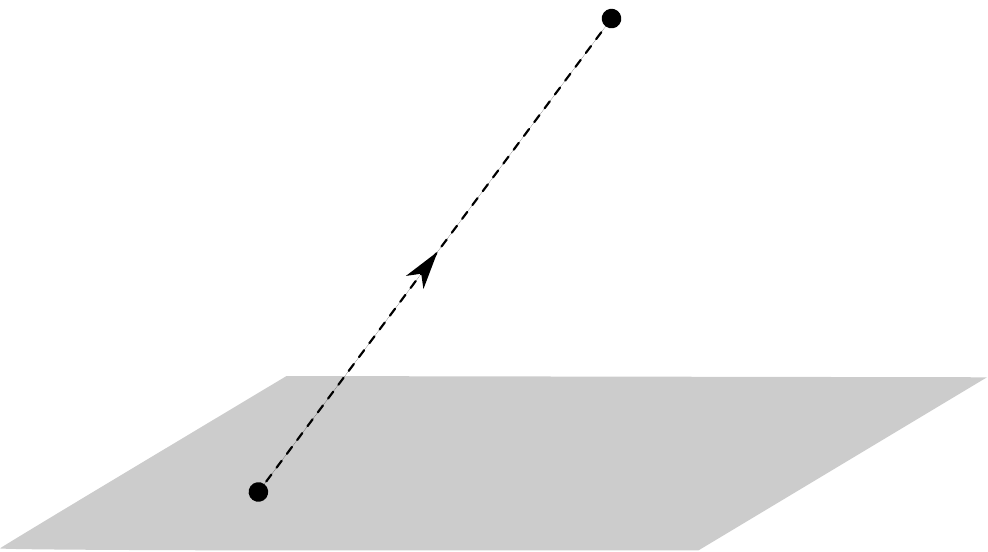}}%
    \put(0.64975137,0.50982245){\color[rgb]{0,0,0}\makebox(0,0)[lt]{\lineheight{1.25}\smash{\begin{tabular}[t]{l}$u$\end{tabular}}}}%
    \put(0.27456852,0.0111617){\color[rgb]{0,0,0}\makebox(0,0)[lt]{\lineheight{1.25}\smash{\begin{tabular}[t]{l}$u_\theta$\end{tabular}}}}%
    \put(0.77610401,0.11035238){\color[rgb]{0,0,0}\makebox(0,0)[lt]{\lineheight{1.25}\smash{\begin{tabular}[t]{l}$\mathbb{R}^{n_\text{output}}$\end{tabular}}}}%
  \end{picture}%
\endgroup%

\def\svgwidth{0.49\textwidth}
\begingroup%
  \makeatletter%
  \providecommand\color[2][]{%
    \errmessage{(Inkscape) Color is used for the text in Inkscape, but the package 'color.sty' is not loaded}%
    \renewcommand\color[2][]{}%
  }%
  \providecommand\transparent[1]{%
    \errmessage{(Inkscape) Transparency is used (non-zero) for the text in Inkscape, but the package 'transparent.sty' is not loaded}%
    \renewcommand\transparent[1]{}%
  }%
  \providecommand\rotatebox[2]{#2}%
  \newcommand*\fsize{\dimexpr\f@size pt\relax}%
  \newcommand*\lineheight[1]{\fontsize{\fsize}{#1\fsize}\selectfont}%
  \ifx\svgwidth\undefined%
    \setlength{\unitlength}{473.76901567bp}%
    \ifx\svgscale\undefined%
      \relax%
    \else%
      \setlength{\unitlength}{\unitlength * \real{\svgscale}}%
    \fi%
  \else%
    \setlength{\unitlength}{\svgwidth}%
  \fi%
  \global\let\svgwidth\undefined%
  \global\let\svgscale\undefined%
  \makeatother%
  \begin{picture}(1,0.55950902)%
    \lineheight{1}%
    \setlength\tabcolsep{0pt}%
    \put(0,0){\includegraphics[width=\unitlength,page=1]{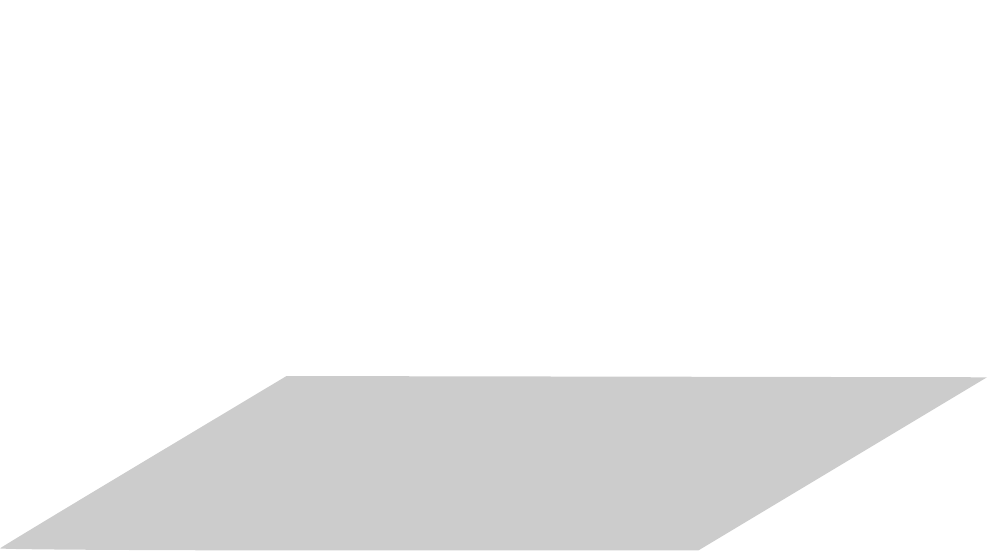}}%
    \put(0.64975137,0.51078092){\color[rgb]{0,0,0}\makebox(0,0)[lt]{\lineheight{1.25}\smash{\begin{tabular}[t]{l}$u$\end{tabular}}}}%
    \put(0.27456852,0.01212019){\color[rgb]{0,0,0}\makebox(0,0)[lt]{\lineheight{1.25}\smash{\begin{tabular}[t]{l}$u_{h, \theta}$\end{tabular}}}}%
    \put(0.77545775,0.10809614){\color[rgb]{0,0,0}\makebox(0,0)[lt]{\lineheight{1.25}\smash{\begin{tabular}[t]{l}$V_h$\end{tabular}}}}%
    \put(0,0){\includegraphics[width=\unitlength,page=2]{discrete_network_output_fem.pdf}}%
    \put(0.62283947,0.07544218){\color[rgb]{0,0,0}\makebox(0,0)[lt]{\lineheight{1.25}\smash{\begin{tabular}[t]{l}$u_h$\end{tabular}}}}%
  \end{picture}%
\endgroup%

\caption{\textbf{Left:} Simplified default approach to learning PDE-solution $u$. \textbf{Right:} Approach of instead learning the corresponding finite element solution $u_h$.}
\label{fig_discnetout_deffem}
\end{figure}

The general idea behind learning finite element solution operators is that regardless of machine learning method, a predicted PDE solution will be on a discrete form. It might therefore be reasonable to consider a discrete form related to a well-studied established method, rather than some seemingly arbitrary form specific to the learning approach. A motivation being that the established method might piggyback the machine learning one to enhance the final product. The established method we consider here is FEM due to its ubiquity, flexibilty, and substantial theory, but the general idea may be applied to other numerical methods together with other machine learning frameworks. See Figure~\ref{fig_discnetout_deffem} for an illustration representing this general idea, where $n_\text{output}$ denotes the output dimension of an arbitrary discrete form of the predicted PDE solution. The operator networks we consider thus map into a finite element space $V_h$ which allows this machine learning approach for PDEs to smoothly be combined with the powerful finite element method. This has a couple of advantages. Firstly, in analyzing the approximation error of the learned solution, tools from finite element theory may be incorporated to provide better error estimates. Secondly, existing finite element software may be used to improve the learned results. In this work we provide examples of both of these advantages.

Furthermore, we provide a naturally parallelizable algorithm for computing the energy in each iteration, which is necessary for efficient training on GPUs. We also suggest to use random batches of elements in the mesh, which has potential to be useful in large problems with many degrees of freedom.

The paper is organized as follows. In Section~\ref{sec:general} we present the general setting by introducing abstract spaces and energy functionals. In Section~\ref{sec:learning} we describe the neural network approach with energy-based loss functions, present some error analysis, and discuss the decomposition of the energy functional for efficient learning. Finally, in Section~\ref{sec:examples} we present several numerical examples where operator learning turns out to be useful.

\section{General setting and energy functionals}\label{sec:general}
Consider a general nonlinear PDE which admits an energy functional $E: V \to \IR$ for a suitable space $V$. We are particularly interested in settings where the energy functional may depend on a parameter $\bfp \in D \subseteq \IR^d$, typically through boundary conditions, coefficients, or right-hand sides. For a given set of parameters $\bfp$, the solution $u \in V$ to the PDE is given by the minimization problem
\begin{align}
	u = \argmin_{v \in V} E(v) \label{min_problem}
\end{align}
where we emphasize that $E(v)$ depends on $\bfp$.

In this paper, we are interested in the solution operator to \eqref{min_problem}. Let $\mcA:D \to V$ denote this solution operator such that $u=\mcA(\bfp)$. Here $\bfp$ is referred to as the problem data variable(s), or simply the parameter(s), and $D$ the parameter space. Furthermore, we assume that $\bfp$ is a random variable with a given probability distribution $\mcP$ over $D$.

\subsection{Finite element discretization}
To be able to learn the solution operator we first need to discretize $V$. For this purpose, we introduce a finite element discretization of the solution space $V$. Let $\mcT_h$ denote a mesh based on a finite subdivision of $\Omega$ into closed and convex elements of maximal diameter $h$, i.e., $\mathrm{diam}(T) \leq h$ for $T\in \mcT_h$. Let $V_h \subseteq V$ be suitable finite element space based on the mesh $\mcT_h$. The corresponding approximation $u_h \in V_h$ is attained by minimizing the energy functional over the finite element space $V_h$
\begin{align}
	u_h = \argmin_{v \in V_h} E(v) \label{min_problem_fem}
\end{align}
We now define the discrete version of $\mcA$,
\begin{align}
	\mcA_h:D \to V_h \quad \text{such that} \quad u_h = \mcA_h(\bfp) \label{operator_fem}
\end{align}
The aim is to set up a neural network that approximates the mapping $\mcA_h$.

\begin{rem}
	We emphasize that the definition of $\mcT_h$ and $V_h$ are intentionally very general. The numerical examples in Section~\ref{sec:examples} will be based on first order Lagrangian (P1) finite elements on triangles. However, in the current framework, it is possible to use many other types of elements and also higher order degree polynomials.
\end{rem}

\subsection{Benefits and limitations}
We remark that not all PDEs have a corresponding energy functional. For those that do not, one could instead use the closely related final element \emph{residual} during training, an example of weak residual minimization which is considered in, e.g., \cite{varnet_2020, berrone_2022}. However, if available, the energy may be preferred to the weak residual. This is because the energy functional generally is computationally cheaper to assemble and that it decomposes more easily into its local contributions, something that we exploit in Section~\ref{sec_efflearn}. Using the weak residual requires assembling the entire system matrix (the stiffness matrix in the Poisson case) since the basis functions of trial and test functions corresponding to neighboring DOFs (mesh nodes for P1 elements) interact within elements to contribute to nonzero entries of the system matrix. This is not the case with the energy where there is only one function, the trial function, which makes the local contributions more isolated.

Drawbacks of using an energy functional are: firstly, not all PDEs have one, as already mentioned; and secondly, that the energy can sometimes be difficult to minimize accurately. The latter is the case in \cite{muller23b}, where a Ritz energy minimization requires finer quadrature for the integration compared to some examples using PINNs strong residual minimizaton. A possible explanation for the increased difficulty could be a flat energy landscape close to the minimum.

A second limitation of our method is that the mesh is hardcoded into the network architecture via the output layer. This means that if one wants to use a different mesh, e.g., by performing refinements or coarsening, the network has to be retrained.

\section{Operator learning with energy minimization}\label{sec:learning}

\subsection{Network architecture and loss function}\label{subsec:archloss}
We consider a simple feedforward and fully connected network, often referred to as multilayer perceptron. This choice of architecture is partly made since, to the best of our knowledge, no other work exists that combines the various components (MLP-approximated finite element solution operator, energy-based loss function, etc.) as is done here. Another argument for choosing MLPs is that we think it is reasonable to first study and explore the basic standard architecture in our setting before considering something more advanced. Although basic, MLPs can posses good approximation capabilities as first presented in \cite{cybenko_1989, hornik_multilayer_1989} for certain types of bounded activation functions, then expanded to any nonconstant bounded function in \cite{HORNIK1991251}, and finally generalized to any nonpolynomial activation in \cite{LESHNO1993861}. This further motivates the architectural choice.

The input layer of the network is the problem data variable $\bfp \in D$ and the output layer is the DOFs of a finite element function in $V_h$. The size of the input layer is $d$, since $\bfp \subseteq \IR^d$, and the size of the output layer equals the number of DOFs, which for P1 elements are the free nodes $n$ in the mesh $\mcT_h$, i.e., nodes that do not correspond to a Dirichlet boundary condition.

The activation function mapping in the hidden layers is set to the exponential linear unit (ELU) function with parameter $\alpha = 1$
\begin{align}
	\sigma(x) = \begin{cases}
	x \quad &\text{if } x >0 \\
	e^x - 1 	&\text{if } x \leq 0
	\end{cases}
\end{align}
This choice was made by testing a few common activations for our examples and then picking the one that worked the best. We point out that ELU should not be viewed as critical for the method, although some of its properties might be. In contrast to the popular and similar rectified linear unit (ReLU), ELU does not have a kink but is instead smooth in the transition from the constant to the increasing part. This smoothness could be relevant for the method since we want the PDE-solution (network output) to depend smoothly on the problem data (network input). It could therefore be good to keep other smooth activations, e.g., swish, in mind when applying the method.

Let $\theta$ denote the trainable parameters (weights and biases) in the network and let
\begin{align}
	\mcA_{h,\theta}:D \to V_h \quad \text{such that} \quad u_{h,\theta} = \mcA_{h,\theta}(\bfp)
\end{align}
describe the neural network approximation of $\mcA_{h}$ in \eqref{operator_fem}.

To train the network we use the the following energy-based loss function
\begin{align}
	\mcL(\theta) := \mathbb{E}_{\bfp \sim \mcP} \left[ E(\mcA_{h,\theta}(\bfp))  \right] \label{loss_fcn_full}
\end{align}
The network thus learns the operator by minimizing the expected value of the energy with respect to the parameter $\bfp$. Since minimizing the energy (for a fixed $\bfp$) is equivalent to solving the finite element problem, we expect the network to accurately approximate the (discretized) solution operator. This is further motivated in Section~\ref{sec_apperrest}.

\subsection{Approximation error estimate} \label{sec_apperrest}
The method is based on the idea of learning finite element solution operators. This turns out to be beneficial when studying the error of the method. First we observe that during training, the loss function $\mcL(\theta) = \mathbb{E}_{\bfp \sim \mcP} \left[ E(\mcA_{h, \theta}(\bfp)) \right]$ is indirectly minimized over a collection of functions in $V_h$ (``indirectly'' since we actually minimize over $\theta$'s for sampled $\bfp$'s, but any combination of these give a function in $V_h$) and  $\mathbb{E}_{\bfp \sim \mcP} \left[ E(\mcA_{h}(\bfp)) \right]$ is the minimum over the set $V_h$. Hence, by assuming well-trainedness of the network we may assume that the error is small, i.e.,
\begin{align}
	\mathbb{E}_{\bfp \sim \mcP} \left[ E(\mcA_{h,\theta}(\bfp)) - E(\mcA_{h}(\bfp)) \right] \leq \epsilon
	\label{eq_assumpwelltrain}
\end{align}
for some $\epsilon>0$. Second, an estimate of the error between the finite element solution and the exact solution is typically known from the rich finite element theory, see for example \cite{brenner_scott}. We shall combine both of these ideas to obtain an approximation error estimate in Proposition~\ref{prp:energyerror}. Letting $\Omega$ be the solution domain, we denote by $\| \cdot \|$ and $(\cdot, \cdot)$ the $L^2(\Omega)$-norm and inner product, respectively. To quantify the error we assume that the energy takes the form
\begin{align}
	E(v) = \frac{1}{2}(A \nabla v,\nabla v) - (f,v)
	\label{eq_energyexp}
\end{align}
which is the case for a Poisson problem with coefficient $A$ and right-hand side $f$. The coefficient $A$ may be a positive scalar field, i.e., $A \geq \alpha > 0$, or a symmetric positive-definite matrix field. For simplicity and clarity, we only consider the scalar case here, but the theoretical results in this section are obtainable for the matrix case as well, using symmetry and positive definiteness. The right-hand side $f \in L^2(\Omega)$. Here we consider both linear\footnote{\label{fn_linen} Strictly speaking, this energy is also nonlinear since it is quadratic. Here, by ``linear energy'' we actually mean an energy with a corresponding \emph{linear PDE}.} and nonlinear energies by letting $A$ depend on either the coordinates of the solution domain, like $v$, or depend on $v$ itself, respectively. We also assume the that discretization is based on Lagrangian finite elements of order $p$. The error is measured in the $H^1$-seminorm also informally called the $H_0^1$-norm
\begin{align}
	|v|_{H^1} = \|v\|_{H_0^1} = \| \nabla v \|
	\label{eq_H01normdef}
\end{align}
To obtain the approximation error estimate we will need the following results that connect the $H_0^1$-norm with the energy.
\begin{lem}[An identity for linear energy]\label{lem:linenergyident}
Let $E(v) = \frac{1}{2}a(v, v) - l(v)$, where $a(\cdot, \cdot)$ is a symmetric, bounded, and coercive bilinear form on $V_h$ and $l(\cdot)$ is a bounded linear form on the same. Also let $a(u_h, v) = l(v)$ for all $v \in V_h$, and $\| v \|_a^2 := a(v, v)$. Then
	\begin{align}
		\frac{1}{2} \| u_{h} - v \|_{a}^2 = E(v) - E(u_{h}) \quad \forall v \in V_h
	\end{align}
	\begin{proof}
		The right-hand side is
		\begin{equation}
		\begin{aligned}
			E(v) - E(u_{h}) &= \bigg(\frac{1}{2} \| v \|_a^2 - l(v) \bigg) - \bigg(\frac{1}{2} \| u_{h} \|_a^2 - l(u_h)\bigg) \\
			&= \frac{1}{2} \| v \|_a^2 - \frac{1}{2} \| u_{h} \|_a^2 + l(\underbrace{u_{h} - v}_{\in V_h}) \\
			&= \frac{1}{2} \| v \|_a^2 - \frac{1}{2} \| u_{h} \|_a^2 + a(u_{h}, u_{h} - v) \\
			&= \frac{1}{2} \| v \|_a^2 - \frac{1}{2} \| u_{h} \|_a^2 + \| u_{h} \|_a^2 - a(u_{h}, v)\\
			&= \frac{1}{2} \| v \|_a^2 + \frac{1}{2} \| u_{h} \|_a^2 - a(u_{h}, v)\\
			&= \frac{1}{2} \| u_{h} - v \|_a^2
		\end{aligned}
		\end{equation}
		which is the left-hand side.
	\end{proof}
\end{lem}

Under appropriate assumptions, the energy $E$ in \eqref{eq_energyexp} fulfills the requirements in Lemma~\ref{lem:linenergyident} with $\| v \|_a^2 = (A \nabla v, \nabla v)$. Thus applying Lemma~\ref{lem:linenergyident} and noting that $\| v \|_a^2 \geq \alpha \| \nabla v \|^2$ where $A \geq \alpha > 0$, we have
\begin{align}
E(v) - E(u_h) = \frac{1}{2} \| u_h - v \|_a^2 \geq \frac{\alpha}{2} \| \nabla u_h - \nabla v \|^2
\end{align}
We summarize this result in the following corollary. We use the notation $\lesssim$ to mean that there is a positive constant involved as a factor on one of the sides (typically the right-hand side) of an inequality $\leq$.

\begin{cor}[An inequality for linear energy]\label{cor:linenergyineq}
Let $E$ be defined by \eqref{eq_energyexp} with $A \geq \alpha > 0$ and $f \in L^2(\Omega)$. Then
\begin{align}
\| \nabla u_{h} - \nabla v \|_{L^2(\Omega)}^2 \lesssim E(v) - E(u_{h}) \quad \forall v \in V_h
\end{align}
\end{cor}

We now consider nonlinear energies by letting the coefficient $A$ depend on the function $v$, i.e., $A = A(v)$. We apply the same approach as in the linear case to relate the difference in energies to the difference in functions. This generalizes the linear result to a nonlinear setting where extra considerations need to be taken into account. We present this as the following lemma.

\begin{lem}[An identity for nonlinear energy]\label{lem:nonlinenergyident}
Let $E(v) = \frac{1}{2}(A(v) \nabla v,\nabla v) - (f, v)$ with $A(v) \geq \alpha > 0$ and $\int_{\Omega} (A(v) + \frac{1}{2}A'(v)v) |\nabla v|^2 \dx > 0$ for all $v \in V_h$. Also let
\begin{equation}
(A(u_h) \nabla u_h, \nabla v) + \frac{1}{2}(A'(u_h)v \nabla u_h, \nabla u_h) = (f, v) \quad \forall v \in V_h \label{nlei_vf}
\end{equation}
and $\| v \|_{A(w)}^2 := (A(w) \nabla v, \nabla v)$. Then
	\begin{align}
	\boxed{
		\frac{1}{2}\| u_h - v \|_{A(u_h)}^2 + R(v, u_h) = E(v) - E(u_{h}) \quad \forall v \in V_h \label{nlei_mainres}
		}
	\end{align}
	where the remainder $R$ is
	\begin{align}
	R(v, u_h) = \frac{1}{2} \int_{\Omega} \big(A(v) - A(u_h)\big) | \nabla v |^2 - A'(u_h) (v - u_h) | \nabla u_h |^2 \dx \label{nlei_remainder}
	\end{align}
	\begin{proof}
		The right-hand side is
		\begin{equation}
		\begin{aligned}
			& \, E(v) - E(u_{h}) \\
			= & \, \bigg( \frac{1}{2}\| v \|_{A(v)}^2 - (f,v) \bigg) - \bigg( \frac{1}{2}\| u_h \|_{A(u_h)}^2 - (f,u_h) \bigg) \\
			= & \, \frac{1}{2}\| v \|_{A(v)}^2 - \frac{1}{2}\| u_h \|_{A(u_h)}^2 + (f,\underbrace{u_h - v}_{\in V_h}) \\
			= & \, \frac{1}{2}\| v \|_{A(v)}^2 - \frac{1}{2}\| u_h \|_{A(u_h)}^2 \\
			& + (A(u_h) \nabla u_h, \nabla (u_h - v)) + \frac{1}{2}(A'(u_h) (u_h - v) \nabla u_h, \nabla u_h) \\
		\end{aligned}
		\end{equation}
		where we have used \eqref{nlei_vf} with test function $u_h - v$ in the last step. Splitting the first term in the last row and adding and subtracting $\frac{1}{2}\| v \|_{A(u_h)}^2$, we have
		\begin{equation}
		\begin{aligned}
			& \, E(v) - E(u_{h}) \\
			= & \, \frac{1}{2}\| v \|_{A(v)}^2 - \frac{1}{2}\| u_h \|_{A(u_h)}^2 + \| u_h \|_{A(u_h)}^2 - (A(u_h) \nabla u_h, \nabla v) \\
			& + \frac{1}{2}(A'(u_h) (u_h - v) \nabla u_h, \nabla u_h) \pm \frac{1}{2}\| v \|_{A(u_h)}^2 \\
			= & \, \frac{1}{2}\| v \|_{A(u_h)}^2 + \frac{1}{2}\| u_h \|_{A(u_h)}^2 - (A(u_h) \nabla u_h, \nabla v) \\
			& + \frac{1}{2}(A'(u_h) (u_h - v) \nabla u_h, \nabla u_h) + \frac{1}{2}\| v \|_{A(v)}^2 - \frac{1}{2}\| v \|_{A(u_h)}^2 \\
			= & \, \frac{1}{2}\| u_h - v \|_{A(u_h)}^2 + \frac{1}{2}\| v \|_{A(v) - A(u_h)}^2 - \frac{1}{2}(A'(u_h) (v - u_h) \nabla u_h, \nabla u_h)
		\end{aligned}
		\end{equation}
		which is the left-hand side of \eqref{nlei_mainres} with $R(v, u_h)$ on norm-product form.
	\end{proof}
\end{lem}

We note that Lemma~\ref{lem:nonlinenergyident} is indeed a generalization of the linear case with $A$ only depending on the coordinates of the solution domain, since then $A'(v) = 0$. The remainder $R$ in \eqref{nlei_mainres} also vanishes in that case, since $A(v) - A(u_h) = A - A = 0$ and $A'(u_h) = 0$.

\begin{rem}
An important thing to point out is that Lemma~\ref{lem:nonlinenergyident} does not guarantee that $v \rightarrow u_h$ when $E(v) \rightarrow E(u_h)$. This is so since the remainder $R$ given by \eqref{nlei_remainder} in general is not necessarily nonnegative. However, convergence can be obtained in special cases depending on two things: the explicit expression of $A$ and the restriction of $v$ to a subset of $V_h$. This is in accordance to what one would expect for nonlinear energies. In general, how easily convergence is obtained depends on both the severity of the nonlinearity (here given by $A$) and the starting point/initial guess (here given by the restriction on $v$).
\end{rem}

We present an example to demonstrate how the remainder $R$ may be handled to obtain the desired convergence.

\begin{ex} \label{ex_nonlinen_r}
We take $A(v) = 1 + v^2$ which is the nonlinear coefficient used for the Poisson energy in \cite{BURMAN2025118111}. This choice satisfies the assumptions on $A$ in Lemma~\ref{lem:nonlinenergyident} since $A(v) \geq 1$ and $A(v) + \frac{1}{2}A'(v)v = 1 + 2 v^2$ for all $v \in V_h$. With this $A$, the remainder given by \eqref{nlei_remainder} is
\begin{equation}
\begin{aligned}
& \, R(v, u_h) \\
= & \, \frac{1}{2} \int_{\Omega} (v^2 - u_h^2) | \nabla v |^2 - 2 u_h (v - u_h) | \nabla u_h |^2 \dx \\
= & \, \frac{1}{2} \int_{\Omega} (v^2 - u_h^2) ( | \nabla v |^2 - | \nabla u_h |^2 ) + (v^2 - u_h^2 - 2 u_h v + 2 u_h^2) | \nabla u_h |^2 \dx \\
= & \, \frac{1}{2} \int_{\Omega} (v^2 - u_h^2) ( | \nabla v |^2 - | \nabla u_h |^2 ) + (v - u_h)^2 | \nabla u_h |^2 \dx
\end{aligned}
\end{equation}
where we have added and subtracted $(v^2 - u_h^2) | \nabla u_h |^2$ to the integrand to obtain the second equality. In the last row, the second term is OK for any $v \in V_h$ since it is nonnegative. The first term is not. However, by restricting $v$ to, e.g., have the same shape as the solution $u_h$ although with different magnitude we may obtain something useful. For instance, if $u_h$ has a hill shape (a reasonable assumption for some Poisson problems) then $v$ can be a smaller or larger such hill (this includes the zero function). In the case with $v$ having a larger magnitude, we note that both factors in the first term are positive: the first $(v^2 - u_h^2)$ trivially so, the second $(| \nabla v |^2 - | \nabla u_h |^2)$ since ``higher hills'' mean steeper slopes. Conversely, in the case with $v$ having a smaller magnitude, both factors are instead negative. So in both cases, the first term will also be nonnegative and OK. We have thus shown for a special case of $A(v)$ how restrictive assumptions on $v$ may be used to guarantee convergence.
\end{ex}

The relation between energy difference and function difference is essential to obtain an estimate for the approximation error (difference between exact and learned solutions) that consists of a finite element error estimate and an estimate for the learning error. Here we restrict ourselves to the case of linear energy to avoid cluttering the core idea with technical details. This also applies to the corresponding estimate for the finite element error, which in the linear case is a well-known result.

\begin{prop}[An estimate for the approximation error]\label{prp:energyerror}
For parameters $\bfp$ from a probability distribution $\mcP$, let $\mcA$ denote the solution operator to the continuous problem \eqref{min_problem} with linear energy functional \eqref{eq_energyexp}. Given a finite element discretization of this problem with mesh size $h$ and based on Lagrangian elements of polynomial degree $p$, let $\mcA_{h,\theta}$ denote the corresponding neural network approximation of the discrete solution operator $\mcA_h$. If we assume well-trainedness of the network with error bound $\epsilon > 0$, i.e., \eqref{eq_assumpwelltrain}, then
\begin{align}
	\mathbb{E}_{\bfp \sim \mcP} \left[ \| \nabla \mcA(\bfp) - \nabla \mcA_{h,\theta}(\bfp) \|_{L^2(\Omega)}^2 \right] \lesssim h^{2p} \bigg( \mathbb{E}_{\bfp \sim \mcP} \left[ \| D^{p+1} \mcA (\bfp) \|_{L^2(\Omega)}^2 \right] \bigg) + \epsilon
\end{align}
\begin{proof}
Recall that $u = \mcA(\bfp)$ and $u_{h,\theta} = \mcA_{h,\theta}(\bfp)$. Hence,
the argument to the expectation operator on left-hand side is
\begin{equation}
\begin{aligned}
	\| \nabla u - \nabla u_{h,\theta} \|^2 &\lesssim \| \nabla u - \nabla u_{h} \|^2 + \| \nabla u_h - \nabla u_{h,\theta} \|^2  \\
					       &\lesssim h^{2p} \| D^{p+1} u \|^2 + E(u_{h,\theta}) - E(u_{h})
\end{aligned}
\end{equation}
Here we have started with an error split using $\pm u_{h}$. Then we have applied a standard error estimate for the finite element error and Corollary~\ref{cor:linenergyineq}
for the learning error, where we have used that $u_{h,\theta} \in V_h$. Taking the expected value of both sides preserves the inequality. Using linearity of the expected value and the well-trainedness assumption \eqref{eq_assumpwelltrain} of the network gives the desired result.
\end{proof}
\end{prop}

\subsection{Efficient learning}  \label{sec_efflearn}
The network will be trained using an iterative method based on the stochastic gradient descent (SGD), e.g., the Adam optimizer. To train the network efficiently, the computation of the gradient of the loss function in \eqref{loss_fcn_full} needs to be fast. Typically, a mini-batch of instances is used to evaluate the expected value in \eqref{loss_fcn_full}. In each iteration of the optimization algorithm, a sample of size $M$, i.e., $\{\bfp_i\}_{i=1}^M$, is generated from the distribution given by $\mcP$. The loss function in \eqref{loss_fcn_full} is thus approximated by
\begin{align}
	\frac{1}{M}\sum_{i=1}^M E(\mcA_{h,\theta}(\bfp_i)) \label{loss_fcn_batch}
\end{align}

We note that the energy in $\eqref{loss_fcn_batch}$ requires a computation on each mesh element, i.e.,
\begin{align}
	 E(\mcA_{h,\theta}(\bfp_i)) = \sum_{T \in \mcT_h} E_T(\mcA_{h,\theta}(\bfp_i))
\end{align}
where $E_T$ denotes the local energy contribution on the element $T$. For, e.g,. a Poisson problem, $E_T$ would require the local assembly of the stiffness and load terms. See Section~\ref{sec:examples} for further examples in practice. We note that these computations can be done in parallel on, e.g., a GPU. For large problems with many elements, we suggest to also use mini-batches of elements. The mini-batch is based on a uniform selection of $N$ elements $\{T_i\}_{i=1}^N$ in each iteration, which is similar to the use of mini-batches in SGD. This means that the loss function is further approximated by
\begin{align}
	\frac{1}{M}\sum_{i=1}^M \sum_{j=1}^N E_{T_j}(\mcA_{h,\theta}(\bfp_i)) \label{loss_fcn_stoch}
\end{align}
Note that we do not divide by $N$, since we do not aim to approximate an expected value but rather the integral over the full domain $\Omega$. Further, we note that if $N$ is equal to the total number of elements in the mesh, then the expression in \eqref{loss_fcn_stoch} equals \eqref{loss_fcn_batch}.

\section{Examples}\label{sec:examples}
In this section we provide three examples to evaluate how the proposed method performs in practice. The first example considers a relatively simple parameterized PDE to introduce the energy minimization approach. The second example concerns a stochastic PDE with a random coefficient depending on a Gaussian random field. Gaussian random fields are of particular interest in, e.g., geophysical applications where the field can be used to describe porous media. The trained network can be used in uncertainty quantification, where many different solutions to the given PDE, corresponding to different realizations of the field, are needed. The third example concerns an application of learning solution operators for nonlinear elasticity problems. Here, the network output is combined with conventional finite element software (FEniCS) to preserve accuracy of the final result and at the same time speed up the computations. This is done by using the learned solution as an initial guess for Newton's method.

The implementation used for the examples is based on the JAX Python code presented in \cite{sharp2023} which is publicly available at \url{https://github.com/nmwsharp/neural-physics-subspaces}. The GPU resources used in the examples were provided by the National Academic Infrastructure for Supercomputing in Sweden (NAISS). More precisely these resources were NVIDIA Tesla A100 HGX GPUs on the Alvis cluster.

Many learning hyperparameters used in the examples such as network depth and training settings have been largely unchanged from the original work \cite{sharp2023}. Although we have experimented with different values and settings, improving upon the original ones have mainly been unfruitful. To evaluate different hyperparameters we have measured learning errors (difference between network prediction and reference finite element solution) as explained below in the different examples. The data-free learning approach we consider here means that we randomly sample both training and test data from the same probability distributions. Thus no cross-validation has been used.

The networks have all had 5 layers: 4 hidden with ELU activation and 1 output with no activation. See Section~\ref{subsec:archloss} for a discussion about the choice of activation function. Going above or below 5 layers typically worsens network performance in our experience. During experimentation the 4 hidden layers were mainly kept at the same width, forming a ``nonlinear block''. We did test with alternatives such as having successively wider layers from input to output layer, but it did not work as well. The width of the nonlinear block is important, or rather getting it in the right range. Luckily this is quite straightforward. We have seen that when starting from a small block width and then increasing it, e.g., doubling it, there has at some point been a substantial improvement in how the network captures the physics. Increasing the width further does not give that much better results. In fact, if one keeps on increasing the width, the results will get worse, since the larger network with more parameters will be harder to train well. We have only used the standard random initialization of the network parameters provided by the Equinox library.

The training has been performed with the Adam optimization algorithm. The default number of iterations have been $10^6$. This might be unnecessarily many in some cases, but during testing we never saw any improvement when going above this number. In each iteration we sample a mini-batch of network inputs. The standard batch size has been 32. For larger networks, we have observed that it can be beneficial to increase the batch size. The training has also employed a decreasing learning rate. The default setting has been to start with a learning rate of $10^{-4}$ and then halve it after every 250k iterations. We have not seen any noteworthy improvements by changing the learning rate settings.

\subsection{A parameterized PDE}\label{ex:parameter}
Consider the following parameterized Poisson problem
\begin{equation}
\left\{
\begin{alignedat}{2}
	-\nabla \cdot(A(\bfp)  \nabla u) &= f \quad &&\text{in } \Omega \\
	u &= 0 &&\text{on } \partial \Omega
\end{alignedat}
\right.
\label{parameter_PDE}
\end{equation}
with $\bfp = (x_0,y_0,r)$, $f=1$, and
\begin{align}
	A(\bfp,x,y) = A(x_0,y_0,r,x,y) = 0.1 + \exp(-((x-x_0)^2 + (y-y_0)^2)/r)
\end{align}
This is a coefficient which is equal to $0.1$ everywhere, except for a spike given by the exponential term. The parameters $\bfp = (x_0,y_0,r)$ determine the location and the radius of the spike. In this example we consider $\Omega = [-1,1]\times[-1,1]$, a uniform distribution for the coordinates $x_0,y_0 \sim \mcU([-1,1])$ (independent), and another uniform distribution for the radius $r \sim \mcU([0.01,0.2])$.

The weak form of \eqref{parameter_PDE} is given by the following: Find $u \in H^1_0$ such that
\begin{align}
	(A(\bfp)\nabla u, \nabla v)_{L^2} = (f, v)_{L^2} \quad \forall v \in H^1_0
\end{align}
which induces the following energy
\begin{align}
	E(v) = \frac{1}{2}(A(\bfp)\nabla v, \nabla v)_{L^2} - (f, v)_{L^2}
\end{align}
For the discretization of $H^1_0$ we consider classical piecewise linear and continuous finite elements based on a  triangulation $\mcT_h$ of the domain $\Omega$. Let $V_h$ denote this finite element space. Note that for $v \in V_h$ the computation of the energy requires the stiffness and load terms, which can be assembled locally on each triangle in the domain, i.e.,
\begin{align}
	E_T(v) = \frac{1}{2} \int_T A(\bfp)\nabla v \cdot \nabla v \dx - \int_T fv \dx
\end{align}

We consider a uniform mesh of size $h=\sqrt{2}\cdot 2^{-4}$, which results in 961 degrees of freedom. This means that the output layer of the network contains 961 nodes, while the input layer consists of 3 nodes corresponding to $(x_0,y_0,r)$. The network is trained using the Adam optimizer and $10^6$ iterations. We use two different widths to explore the impact on accuracy. The training times are reported in Table~\ref{table_xyr_times}. We deduce that using batches of triangles ($T=32$) has no effect. The training times are roughly the same as for the full mesh. This is due to the fact that energy computation is done in parallel and the problem is relatively small (the GPU Utilization is not 100\%).

\begin{table}[h!]
	\caption{Training and inference time for the parameterized problem in Section~\ref{ex:parameter}. Output layer of size 961. Trained on an A100 GPU.}
	\centering
	\begin{tabular}{|l|l|l|l|}
		\hline
		$T$ & Dim & Training time & Inference time \\ \hline
		$32$ & $4 \cdot 256$ & 458 s &  1 ms\\  \hline
		Full & $4 \cdot 256$ & 433 s &  1 ms\\ \hline
		$32$ & $4 \cdot 512$ & 440 s & 1 ms \\ \hline
		Full & $4 \cdot 512$ & 426 s & 1 ms \\ \hline
	\end{tabular}\label{table_xyr_times}
\end{table}

\begin{figure}[h!]
	\centering
	\begin{subfigure}[b]{0.35\textwidth}
		\includegraphics[width=\textwidth]{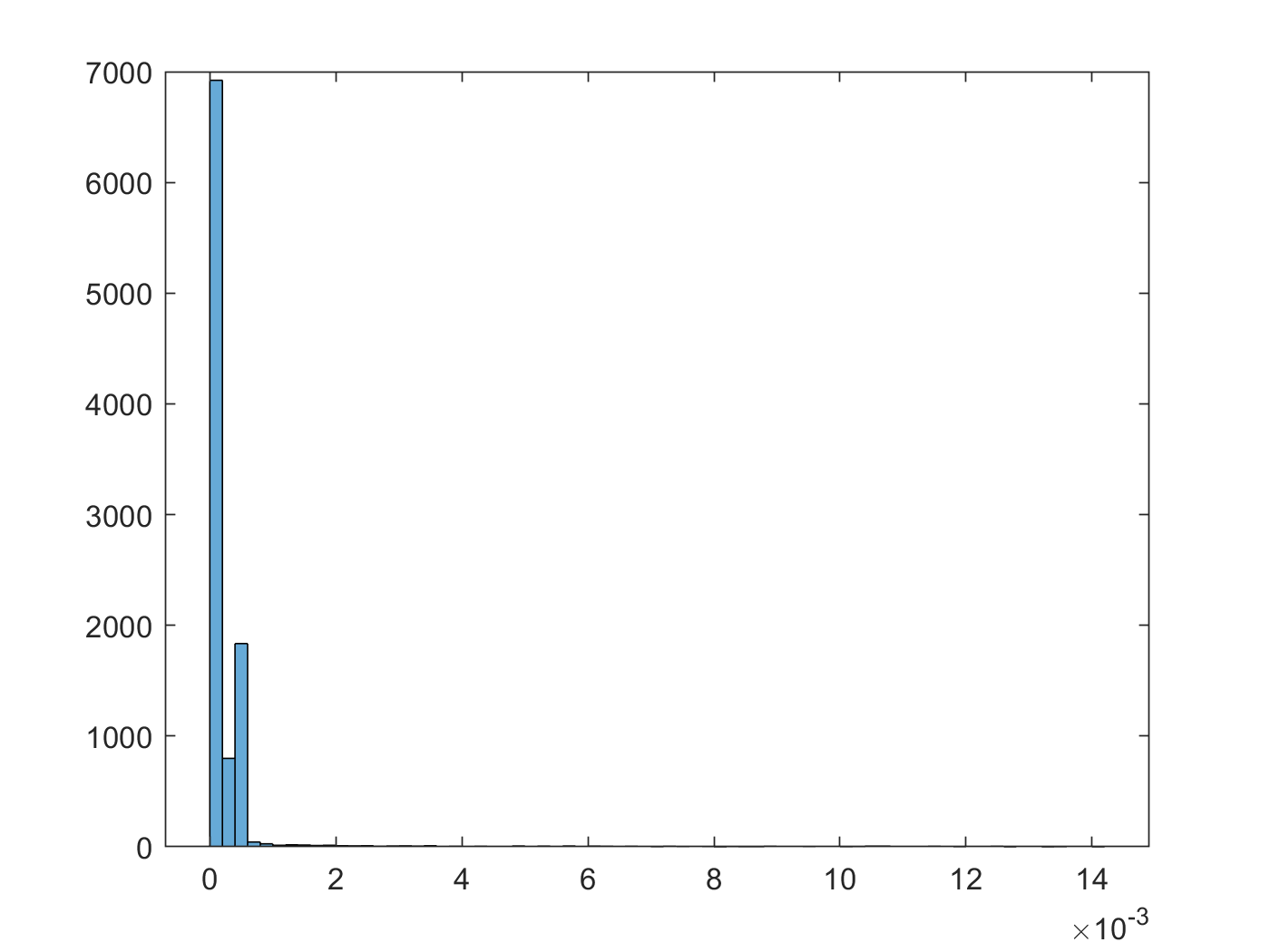}
		\caption{Energy}
	\end{subfigure}\hspace{-1.4em}
	\begin{subfigure}[b]{0.35\textwidth}
		\includegraphics[width=\textwidth]{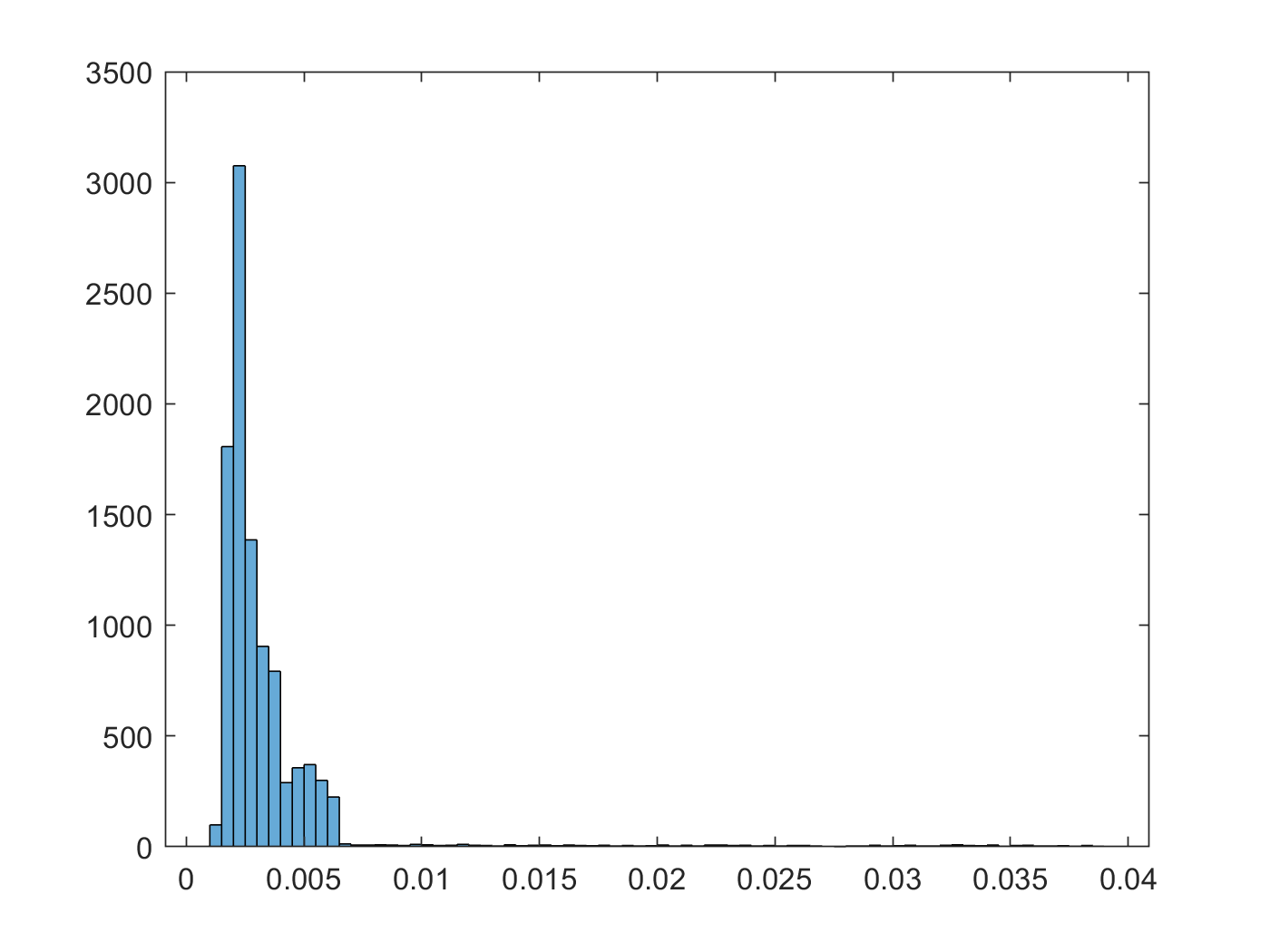}
		\caption{$L^2$-norm}
	\end{subfigure}\hspace{-1.4em}
	\begin{subfigure}[b]{0.35\textwidth}
		\includegraphics[width=\textwidth]{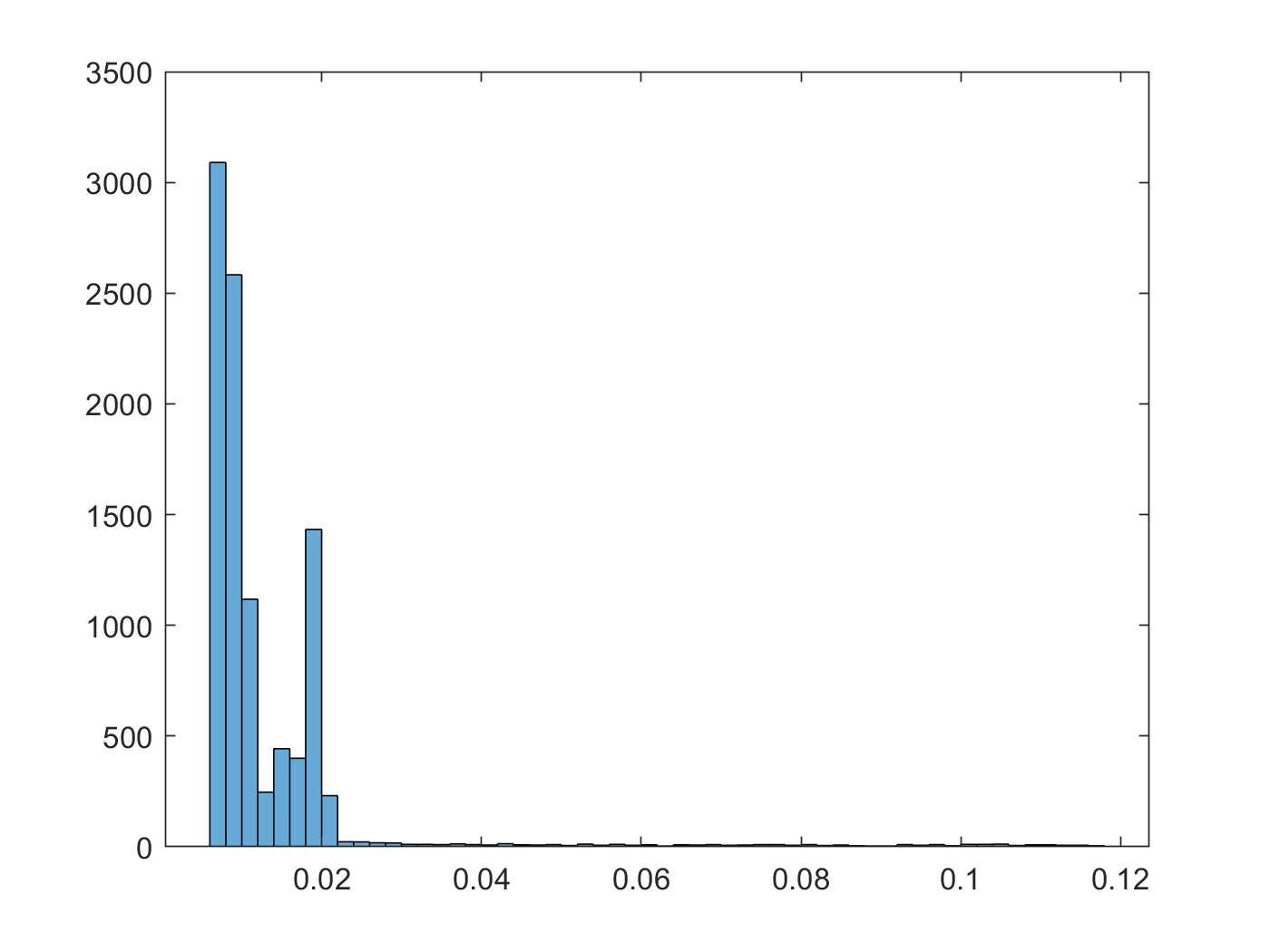}
		\caption{$H^1$-norm}
	\end{subfigure}
	\caption{Histogram plots for the different relative errors for a network of size $4\cdot 256$ using all elements (full) per batch, i.e., the network in the second row of Table~\ref{table_xyr_errors}. Sample size $K = 10^4$.} \label{fig_xyr_histograms}
\end{figure}

To investigate the error of the method we compute (an approximation of) the expected value of the difference between the output of the network and the finite element solution in different norms. Given a sample of parameters $\{\bfp_i \}_{i=1}^K$ we compute the difference in the energy, the $L^2$-norm, and the $H^1$-norm, i.e., we compute the relative errors
\begin{align}
	&|E(\mcA_{h,\theta}(\bfp_i)) - E(\mcA_{h}(\bfp_i))|/|E(\mcA_{h}(\bfp_i))| \label{eq_diffenerg} \\
	&\|\mcA_{h,\theta}(\bfp_i) - \mcA_{h}(\bfp_i)\|_{L^2}/\|\mcA_{h}(\bfp_i)\|_{L^2} \\
	&\|\mcA_{h,\theta}(\bfp_i) - \mcA_{h}(\bfp_i)\|_{H^1}/\|\mcA_{h}(\bfp_i)\|_{H^1}
\end{align}
Note that the difference in the energy \eqref{eq_diffenerg} gives an approximation of the \emph{relative} well-trainedness of the network, cf., \eqref{eq_assumpwelltrain}. In Table~\ref{table_xyr_errors} we report mean and standard deviation of these relative errors computed using $K=10^4$ samples of the parameter $\bfp$. Corresponding histogram plots are shown in Figure~\ref{fig_xyr_histograms} for the network of size $4\cdot 256$ and batches using all elements in the mesh. We note that the error distribution is small which implies that the network is well-trained over the distribution of the parameters. Furthermore, comparing the errors in Table~\ref{table_xyr_errors} we see that increasing the width of the network improves the approximation properties. We also note that the error naturally increases when using smaller batch sizes of triangles.

\begin{table}[h!]
	\caption{Relative errors for the parameterized problem in Section~\ref{ex:parameter}. Output layer of size 961. Sample size $K = 10^4$. Training and test data $x_0,y_0 \sim \mcU([-1,1])$, $r \sim \mcU([0.01,0.2])$.}
	\centering
	{
	\sisetup{scientific-notation=true, output-exponent-marker = e, table-format=1.1e1, table-text-alignment=left}
	\begin{tabular}{| l | l | S | S | S | S | S | S |}
		\hline
		$T$ & Dim & {Energy} & {Std} & {$L^2$-norm} & {Std} & {$H^1$-norm} & {Std} \\ \hline
		$32$ & $4 \cdot 256$ & 0.0061 & 0.0055 & 0.026 & 0.011  & 0.065 & 0.026 \\ \hline
		Full & $4 \cdot 256$ & 4.1e-4 & 0.0013 & 0.0036 & 0.0041 & 0.014 & 0.013 \\ \hline
		$32$ & $4 \cdot 512$ & 0.0045 & 0.0042 & 0.024 & 0.010  & 0.057 & 0.021 \\ \hline
		Full & $4 \cdot 512$ & 2.0e-4 & 7.3e-4 & 0.0020 & 0.0022 & 0.0092  & 0.0093 \\ \hline
	\end{tabular}\label{table_xyr_errors}
	}
\end{table}

To show a case where batches of triangles do have an effect on the training times, we consider a finer mesh of size $h=\sqrt{2}\cdot 2^{-6}$ which results in 16,129 degrees of freedom and 32,768 triangles. We also increase the width of the neural network to $4\cdot 2048$. In Table~\ref{table_xyr_2048_manydofs}--\ref{table_xyr_2048_manydofs_errors} the results are reported for batches of size 3277 ($10\%$ of the elements) compared to the full mesh. We observe that the training time is reduced by using a smaller batch of triangles,
since the GPU Utilization is at 100\% for both cases and the energy computation can not be completely parallelized in each iteration.
The learning error increases though, since the number of iterations is the same in both cases. We emphasize that for large problems and the choice of GPU model, the batches can have significant impact on the computational times. On a laptop GPU, this could also be crucial to be able to train the network at all. The scalability of the method is further explored in the applied paper \cite{BURMAN2025118111} (See Tables~1 and 2 there), where it is used for solving inverse problems of a more complex nature (nonlinear Poisson energy functional with coefficient $A(v) = 1 + v^2$). In this case, the training becomes practically infeasible unless mini-batches are used for the triangles, at least for the smaller mesh sizes.

\begin{table}[h!]
	\caption{A neural network of $4\cdot2048$ hidden layers for the parameterized problem in Section~\ref{ex:parameter}. Output layer of size 16,129. Trained on an A100 GPU.}
\centering
\begin{tabular}{|l|l|l|}
	\hline
	$T$ &  Training time & Inference time \\ \hline
	$3277$ & 1720 s & 1 ms\\ \hline
	Full & 1962 s & 1 ms \\ \hline
\end{tabular}\label{table_xyr_2048_manydofs}
\end{table}

\begin{table}[h!]
	\caption{Relative errors for the parameterized problem in Section~\ref{ex:parameter}. Output layer of size 16,129. Sample size $K = 10^4$. Training and test data $x_0,y_0 \sim \mcU([-1,1])$, $r \sim \mcU([0.01,0.2])$.}
	\centering
	{
	\sisetup{scientific-notation=true, output-exponent-marker = e, table-format=1.1e1, table-text-alignment=left}
	\begin{tabular}{| l | l | S | S | S | S | S | S |}
		\hline
		$T$ & Dim & {Energy} & {Std} & {$L^2$-norm} & {Std} &  {$H^1$-norm} & {Std} \\ \hline
		$3277$ & $4 \cdot 2048$ & 0.0033 & 0.0014 & 0.0090 & 0.0042 & 0.050  & 0.010 \\ \hline
		Full & $4 \cdot 2048$ & 1.2e-04 & 2.7e-04 & 0.0026 & 0.0013 & 0.0082 & 0.0050 \\ \hline
	\end{tabular}\label{table_xyr_2048_manydofs_errors}
	}
\end{table}

We end this numerical example by noting two things: Firstly we note that the inference time, i.e., the time for one forward pass through the network is very small, around 1 ms in all cases studied. This shows the strength of learning the solution operator. Once trained, the solution for a given parameter is obtained using very little effort. Secondly we note that the relative energy learning error \eqref{eq_diffenerg} (``Energy'' in Table~\ref{table_xyr_errors} and \ref{table_xyr_2048_manydofs_errors}) decreases with increased network width, both for hidden layers and output layer (an increase here means a decrease in the mesh size $h$). This lends credence to the well-trainedness assumption \eqref{eq_assumpwelltrain}, and thus to Proposition~\ref{prp:energyerror}, where it is used.

\subsection{Gaussian random fields}\label{ex:grf}
We consider a Poisson problem with a random coefficient based on a Gaussian random field
\begin{equation}
\left\{
\begin{alignedat}{2}
	-\nabla\cdot (A(\omega)  \nabla u) &= f \quad &&\text{in } \Omega \\
	u &= 0 &&\text{on } \partial \Omega
\end{alignedat}
\right.
\end{equation}
where $A(\omega):=\exp{a(\omega)}$ where $a(\omega)$ is a Gaussian random field with covariance function given by
\begin{align}
	C(x,y) = \exp{\bigg(-\frac{|x-y|^2}{2L^2}\bigg)}
\end{align}
where $L$ is the correlation length scale. We use a Karhunen-Lo\`{e}ve expansion to approximate the field
\begin{align}
	a(x,y,\omega) \approx \mu + \sigma \sum_{i=1}^{N_{KL}} \sqrt{\lambda_i}\phi_i  \xi_i(\omega)
\end{align}
where $\mu$ and $\sigma$ are the constant mean and standard deviation of the field, respectively, $\lambda_i$ and $\phi_i$ are the eigenvalues and eigenfunctions corresponding to the covariance, respectively, and $\xi_i \sim \mcN(0,1)$ are i.i.d random variables. The number $N_{KL}$ determines the number of terms in the truncated sum. Similarly to the Poisson problem in Section~\ref{ex:parameter}, the energy in this setting is given by
\begin{align}
	E(v) = \frac{1}{2}(A(\bfxi)\nabla v, \nabla v)_{L^2} - (f, v)_{L^2}
\end{align}
where $\bfxi(\omega) = (\xi_1(\omega),...,\xi_{N_{KL}}(\omega))$. The vector $\bfxi$ takes different values depending on the realization $\omega$ and is thus considered the parameter in this case. For the local computations on a triangle we have, c.f., Section~\ref{ex:parameter},
\begin{align}
	E_T(v) = \frac{1}{2} \int_T A(\bfxi)\nabla v \cdot \nabla v \dx - \int_T fv \dx
\end{align}

\begin{figure}[h!]
	\centering
	\begin{subfigure}[b]{0.35\textwidth}
		\includegraphics[width=\textwidth]{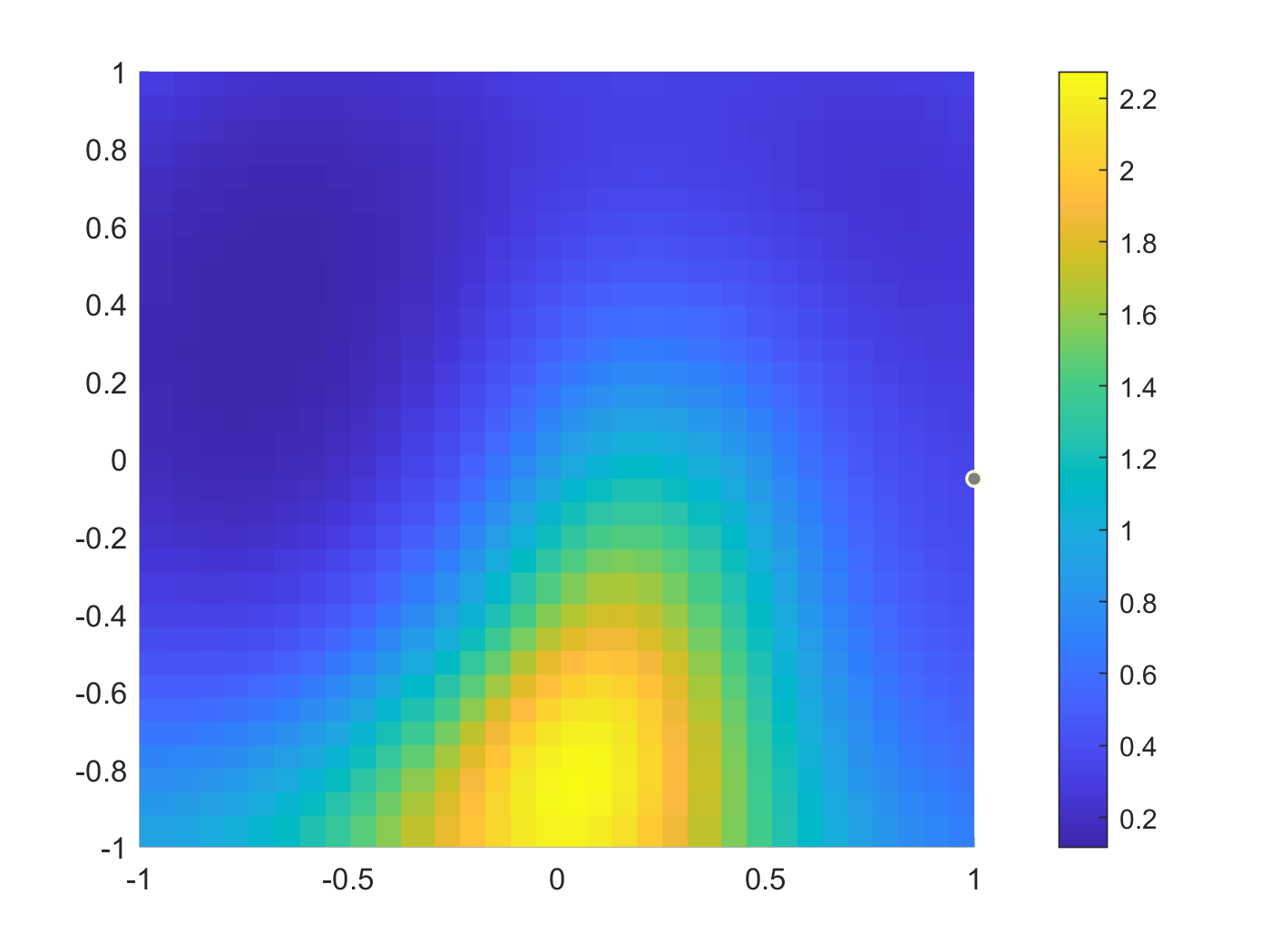}
		\caption{Coefficient}
	\end{subfigure}\hspace{-1.4em}
	\begin{subfigure}[b]{0.35\textwidth}
		\includegraphics[width=\textwidth]{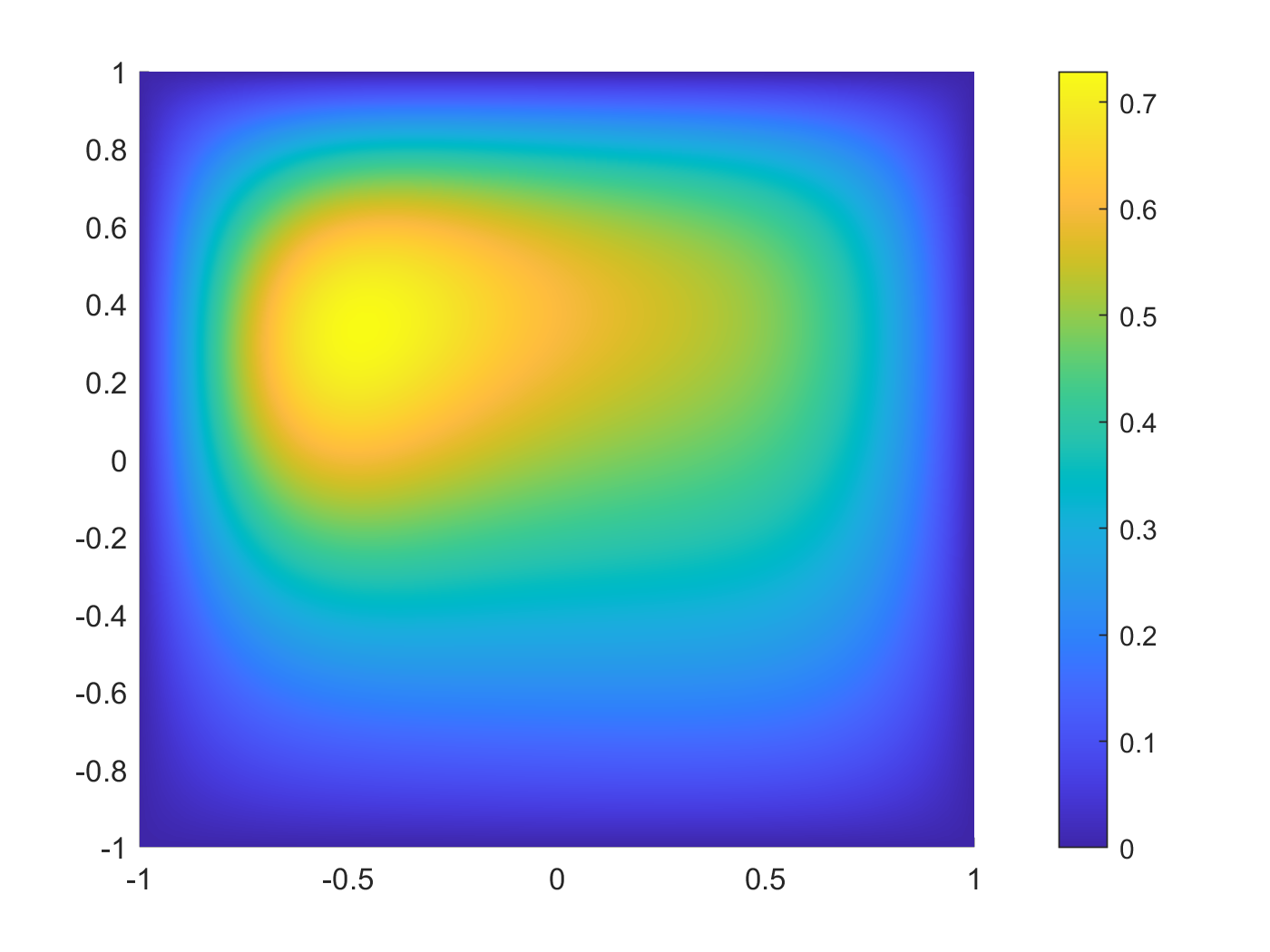}
		\caption{FEM}
	\end{subfigure}\hspace{-1.4em}
	\begin{subfigure}[b]{0.35\textwidth}
		\includegraphics[width=\textwidth]{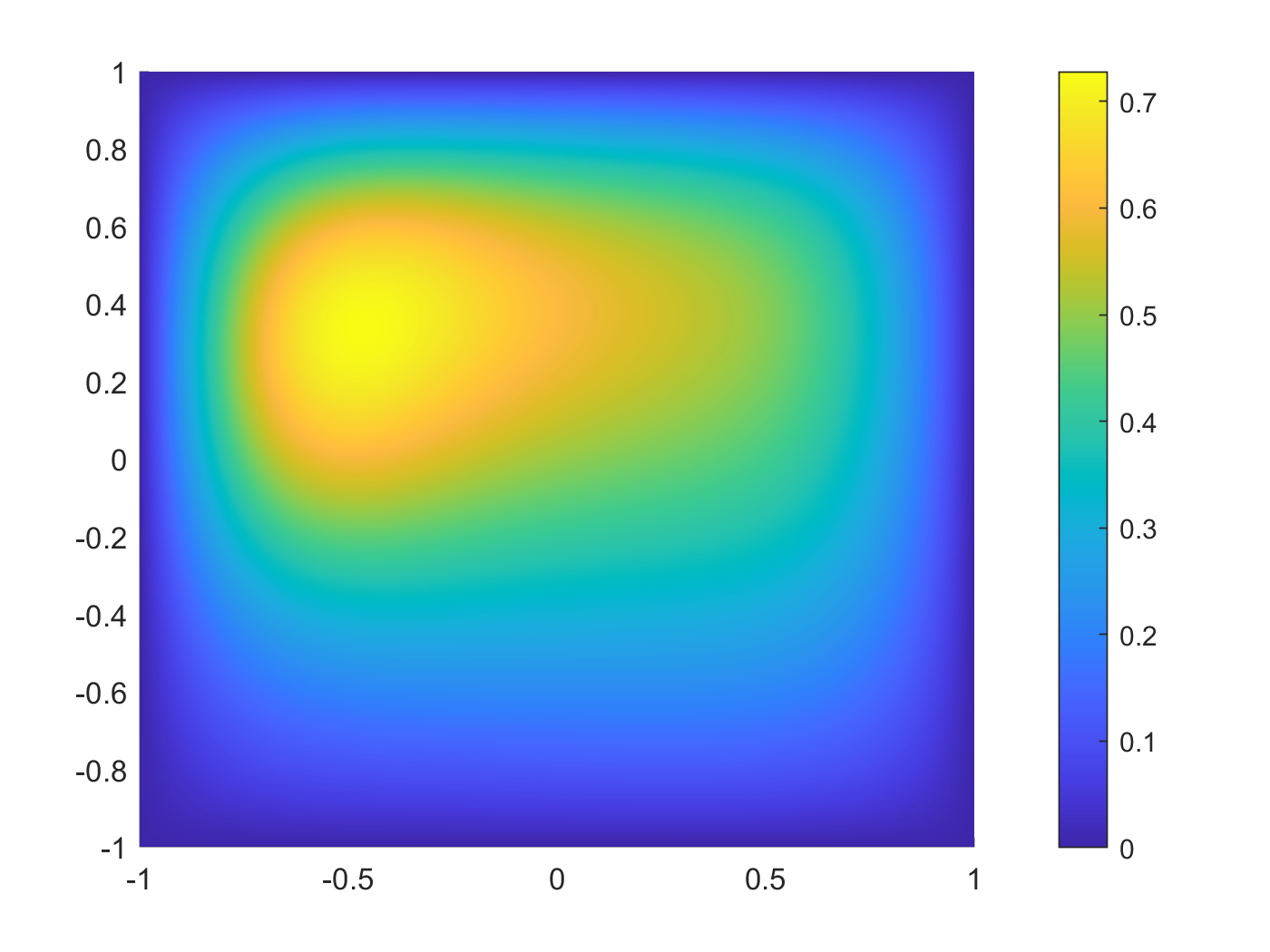}
		\caption{NN}
	\end{subfigure}
	\caption{The figure shows a sample coefficient (a) and the corresponding approximation achieved by FEM (b) and by the trained neural network (c).} \label{fig_grf_example}
\end{figure}

\begin{figure}[h!]
	\centering
	\begin{subfigure}[b]{0.35\textwidth}
		\includegraphics[width=\textwidth]{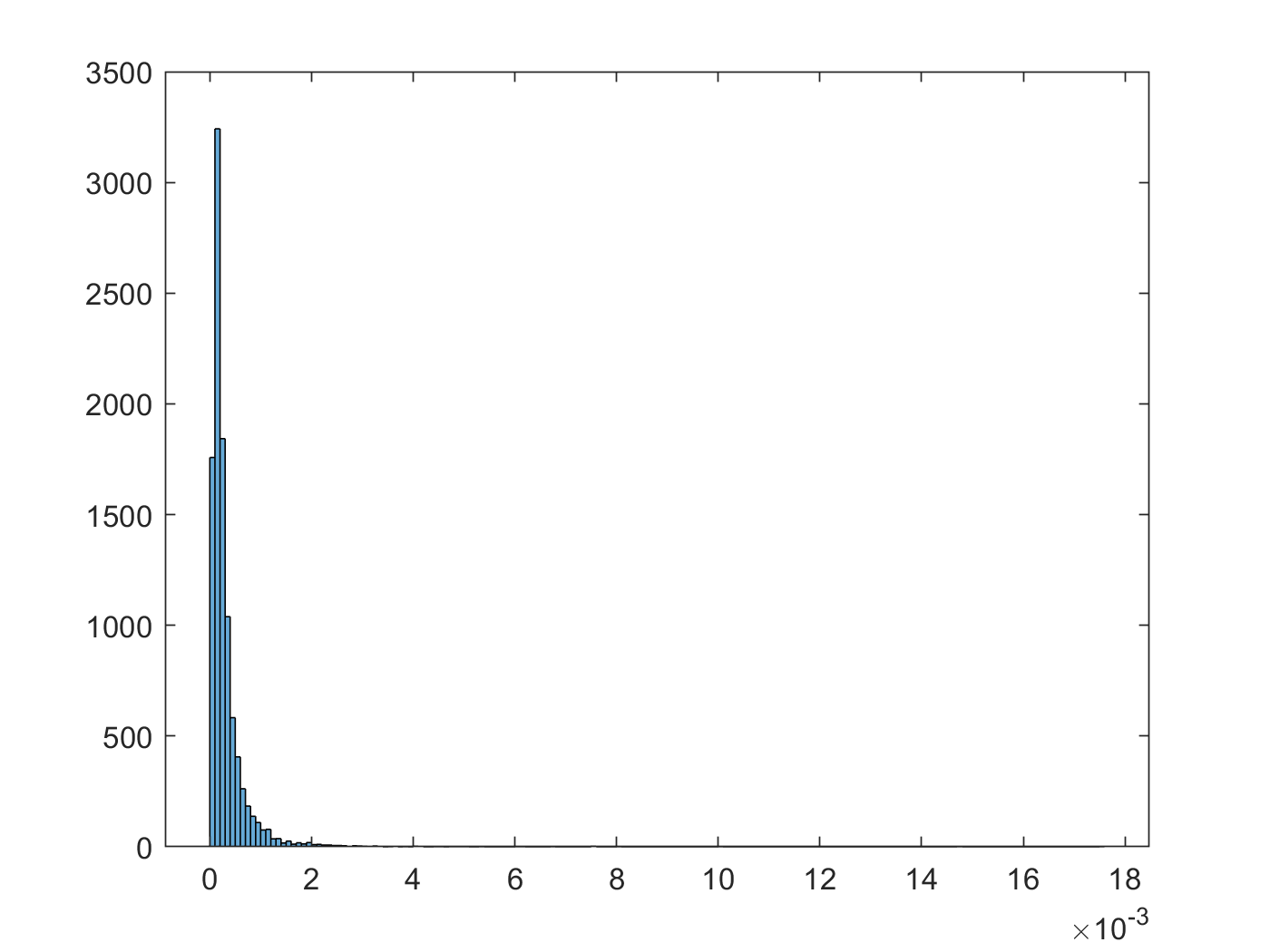}
		\caption{Energy}
	\end{subfigure}\hspace{-1.4em}
	\begin{subfigure}[b]{0.35\textwidth}
		\includegraphics[width=\textwidth]{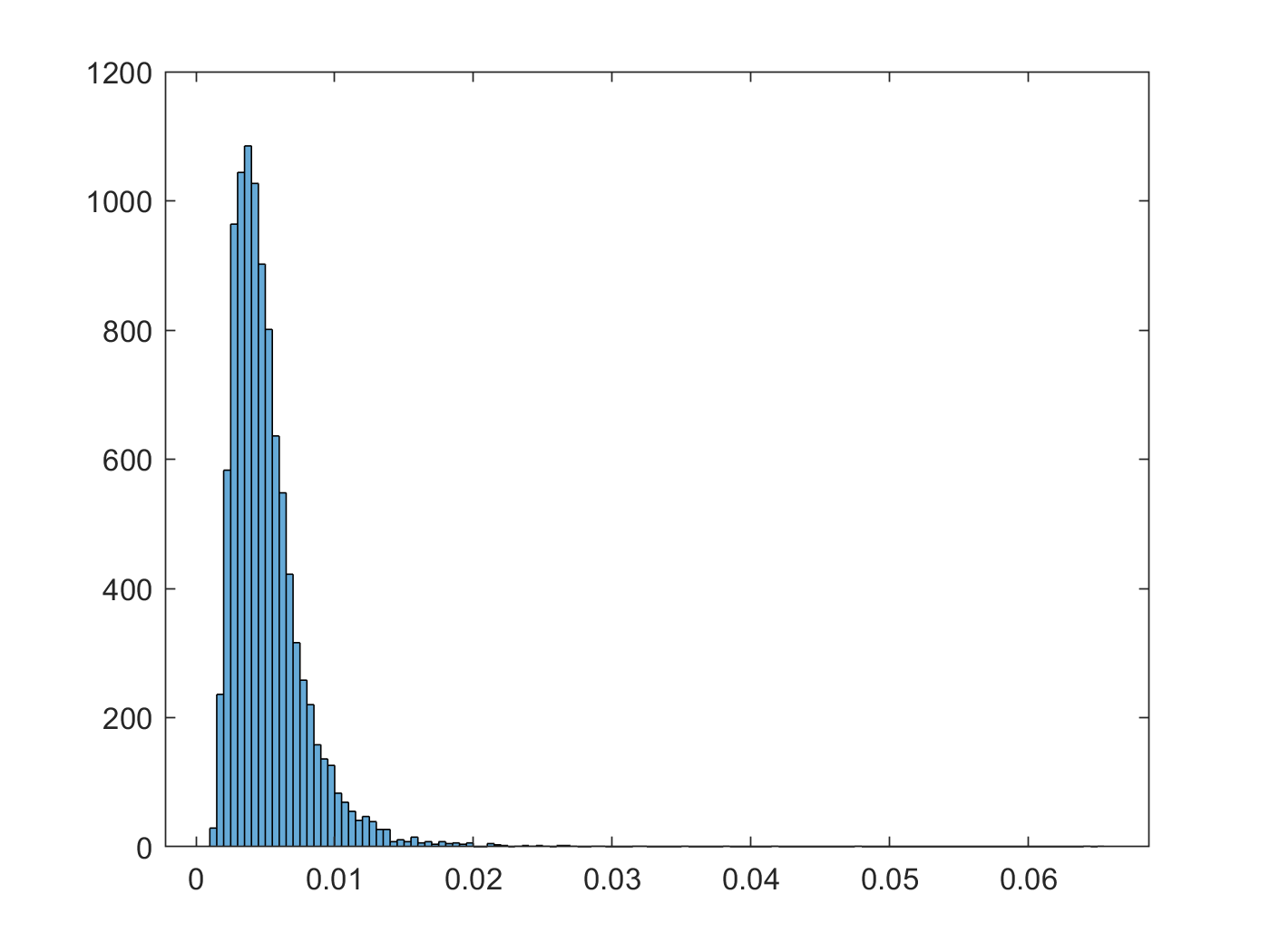}
		\caption{$L^2$-norm}
	\end{subfigure}\hspace{-1.4em}
	\begin{subfigure}[b]{0.35\textwidth}
		\includegraphics[width=\textwidth]{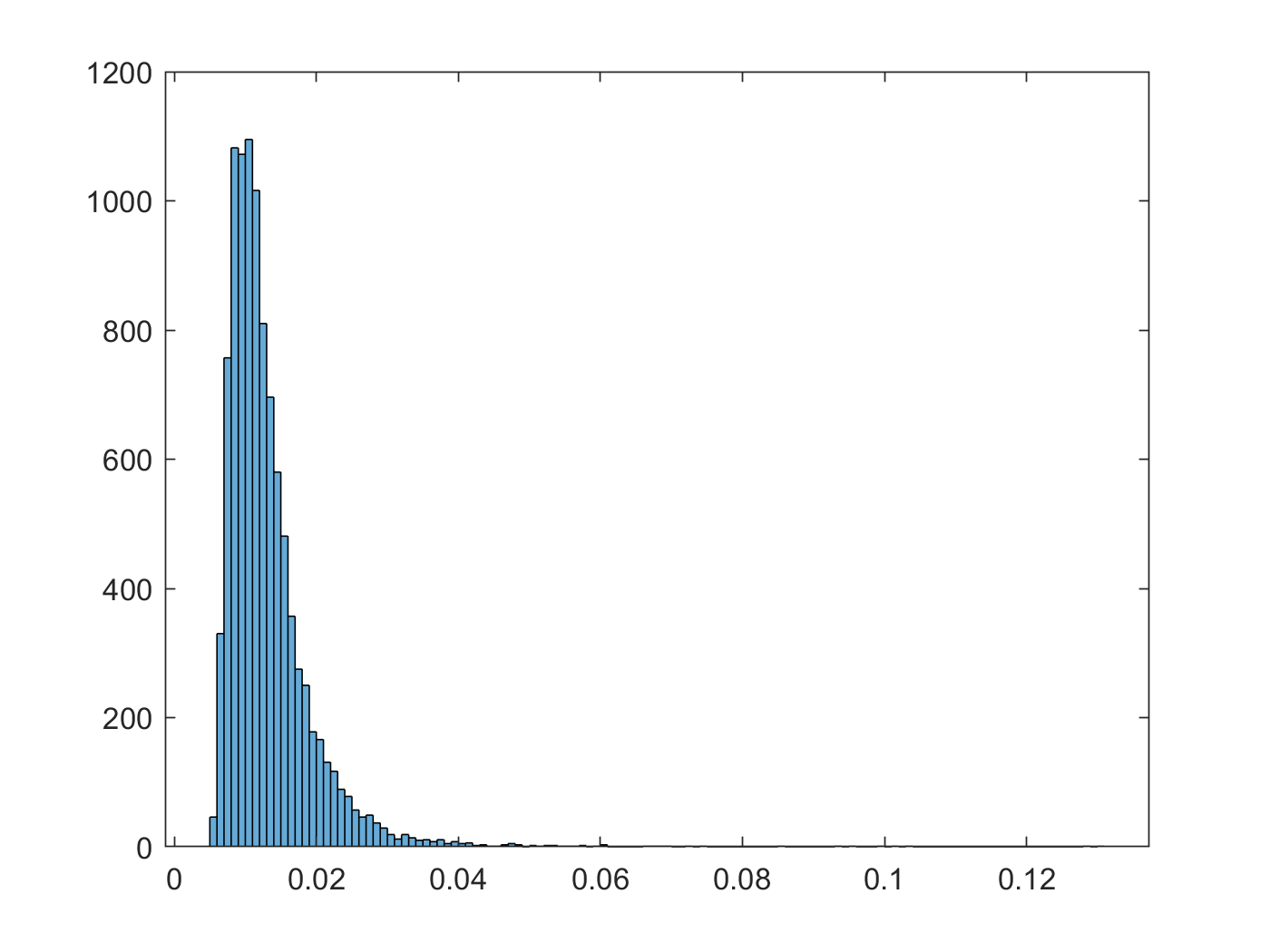}
		\caption{$H^1$-norm}
	\end{subfigure}
	\caption{Histogram plots for the different relative errors for a network of size $4\cdot 512$ using all elements (full) per batch, i.e., the network in Table~\ref{table_xi_512_errors}. Sample size $K = 10^4$.} \label{fig_grf_histograms}
\end{figure}

For the numerical experiment we choose $\mu=0$, $\sigma = 1$, $L=0.5$, and $N_{KL}=9$. The eigenfunctions and eigenvalues are approximated on a Cartesian grid of size $h=2^{-4}$. For the triangulation of the domain, we consider a uniform mesh of size $h=\sqrt{2}\cdot 2^{-5}$, which results in 3969 degrees of freedom. This means that the output layer of the network contains 3969 nodes, while the input layer consists of 9 nodes corresponding to the length of the vector $\bfxi$. We emphasize that given a realization $\omega$, the trained neural network approximates the solution operator by $u_{h,\theta} = \mcA_{h,\theta}(\bfxi(\omega))$. In Table~\ref{table_xi_512_time} we report training time for a $4\cdot512$ network. In Figure~\ref{fig_grf_example} we plot a sample coefficient and the corresponding finite element and neural network approximation. From the plots we can see that the neural network produces an accurate approximation. The error is further analyzed below (see Table~\ref{table_xi_512_errors} and Figure~\ref{fig_grf_histograms}). Note that we do not consider batches of triangles in this example, since the number of degrees of freedom is relatively small.

\begin{table}[h!]
	\caption{A neural network of $4\cdot512$ hidden layers for the Gaussian random field problem in Section~\ref{ex:grf}. The columns show the total training time and inference time. Trained on an A100 GPU.}
	\centering
	\begin{tabular}{|l|l|l|}
		\hline
		T & Training time & Inference time \\ \hline
		Full  & 426 s &  1 ms\\ \hline
	\end{tabular}\label{table_xi_512_time}
\end{table}

Similar to first example in Section~\ref{ex:parameter} we investigate the error by computing an approximation of the expected value of the difference between the output of the network and the finite element solution. That is, given a sample of parameters $\{\bfxi_i \}_{i=1}^K$ we compute the relative errors
\begin{align}
	&|E(\mcA_{h,\theta}(\bfxi_i)) - E(\mcA_{h}(\bfxi_i))|/|E(\mcA_{h}(\bfxi_i))| \\
	&\|\mcA_{h,\theta}(\bfxi_i) - \mcA_{h}(\bfxi_i)\|_{L^2}/\|\mcA_{h}(\bfxi_i)\|_{L^2} \\
	&\|\mcA_{h,\theta}(\bfxi_i) - \mcA_{h}(\bfxi_i)\|_{H^1}/\|\mcA_{h}(\bfxi_i)\|_{H^1}
\end{align}
In Table~\ref{table_xi_512_errors} we report mean and standard deviation of these relative errors computed using $K=10^4$ samples of the parameter $\bfxi$. Corresponding histogram plots are shown in Figure~\ref{fig_grf_histograms}. We note that error distribution is small which implies that the network is well-trained over the distribution of the parameters.

\begin{table}[h!]
	\caption{Relative errors for the Gaussian random field problem. All elements in the mesh are used for a neural network of $4\cdot512$ hidden layers. Sample size $K = 10^4$.}
	\centering
	{
	\sisetup{scientific-notation=true, output-exponent-marker = e, table-format=1.1e1, table-text-alignment=left,  round-mode=places, round-precision=1}
	\begin{tabular}{| l | l | S | S | S |}
		\hline
		Training \& test data & Measure & {Energy} & {$L^2$-norm} & {$H^1$-norm} \\ \hline
		\multirow{2}{*}{$\xi_i \sim \mcN(0, 1) \quad i = 1, \dots, 9$} & Mean & 3.2e-4 & 0.0052& 0.0131 \\ \cline{2-5}
		& Std & 4.8e-4 & 0.0029 & 0.0064\\ \hline
	\end{tabular}\label{table_xi_512_errors}
	}
\end{table}

For stochastic PDEs we are typically interested in a quantity of interest (QoI), for instance the expected value of the $L^2$-norm of the solution
\begin{align}
	\mathbb{E}_{\bfxi \in \mcN(0,1)}\left[\|u(\bfxi)\|^2_{L^2}\right]
	\label{eq_qoi1}
\end{align}
or the average point value at some coordinate $x_0$
\begin{align}
	\mathbb{E}_{\bfxi \in \mcN(0,1)}\left[u(x_0,\bfxi)\right]
	\label{eq_qoi2}
\end{align}
A classical Monte Carlo approach to approximate the quantity will require many evaluations of the solution operator $u = \mcA(\bfxi)$. For the finite element approximation, this means that we would need to compute an approximation for each realization $\omega$. In particular, the stiffness matrix needs to be reassembled for each such realization. However, for the trained neural network, retrieving an approximation for a given realization is a simple forward pass of the network which is typically very fast (around 1 ms). This step is also easily parallelized on the GPU using the JAX framework. To show the advantage of the neural network approach, we compare the computational times for evaluating the quantities of interest \eqref{eq_qoi1} and \eqref{eq_qoi2} using the trained neural network with the corresponding times from using a finite element approach with the same degrees of freedom. The implementation is done using the FEniCSx library \cite{fenicsx}. For a fair comparison with the FEM approach, we also compute the quantity of interest using the (trained) neural network sequentially on the CPU. The GPU computations are performed on the same A100 GPU used for the training and the CPU computations on an Intel(R) Core(TM) i9-11950H 2.60GHz. The results are presented in Tables~\ref{table_qoi} and \ref{table_qoi_point}.

\begin{table}[h!]
	\caption{Times for computing $\mathbb{E}_{\bfxi \in \mcN(0,1)}\left[\|u(\bfxi)\|^2_{L^2}\right]$ using $10^4$ samples.}
	\centering
	\begin{tabular}{|l|l|l|l|l|}
		\hline
		Method & QoI & Training & GPU (parallel)  & CPU (sequential) \\ \hline
		NN & 0.186 & 426 s & 0.21 s & 25 s \\ \hline
		FEM & 0.181 & -- &  -- & 79 s \\ \hline
	\end{tabular}\label{table_qoi}
\end{table}

\begin{table}[h!]
	\caption{Times for computing $\mathbb{E}_{\bfxi \in \mcN(0,1)}\left[u(x_0,\bfxi)\right]$ for $x_0 = (0,0)$ using $10^4$ samples.}
	\centering
	\begin{tabular}{|l|l|l|l|l|}
		\hline
		Method & QoI & Training & GPU (parallel) & CPU (sequential)  \\ \hline
		NN & 0.311 & 426 s & 0.26 s & 20 s\\ \hline
		FEM & 0.310 & -- &-- &71 s \\ \hline
	\end{tabular}\label{table_qoi_point}
\end{table}

We observe that, in terms of inference times, the neural network clearly outperforms the FEM implementation. In particular when the inference is parallelized on the GPU, but also on a classical CPU when the inferences are done sequentially. We note that the training time is large in comparison, but once trained, the same neural network can be used to compute any quantity of interest. Using the data in Table~\ref{table_qoi}, we conclude that the neural network approach is beneficial after approximately 78,889 samples using sequential evaluation (CPU). In Table~\ref{table_qoi_point} the corresponding number is 83,530.

It is worth pointing out that JAX has also been used to speed up traditional FEM-solvers by 10--100 times \cite{XUE2023108802, JAXCPFEM, TOUTLINI2026109866}. In Tables~\ref{table_qoi} and \ref{table_qoi_point}, the speed-ups when going from FEM on a CPU to a neural network on a GPU are 376 and 273, respectively, but there are examples of much greater speed-ups than these. For example, in \cite{WANG2025109783} it is reported that the inference of an instance of the FEM-NN model NNDFEM is 8550 times faster than standard FEM when running on the same hardware! Although the neural network approach can be significantly faster than combining JAX with FEM, the latter produces the exact FE-solution whereas the former only an approximation of it. It might thus be relevant to stay up to date with modern speed-up tools for traditional solvers since they could potentially reduce the impact of neural methods.

\subsection{Nonlinear elasticity} \label{sec_nonlinelast}

A promising application of solution operator networks is to use the output as an initial guess for Newton's method when solving nonlinear problems, see, e.g., \cite{huang_int-deep_2020, odot_deepphysics_2022, aghili_accelerating_2024}.
The network output will always be an approximation whose accuracy is determined by the degree of well-trainedness of the network. One could imagine that in some cases and applications, the desired degree of well-trainedness might be expensive to reach or simply unattainable. Instead applying a finishing touch like Newton iterations to the network output is an alternative that could be cheaper, more reasonable, and that removes any accuracy doubts.
This is a \emph{complementary} way of using machine learning for PDE solving rather than as a complete solver method, i.e., to consider the operator network's prediction the finished product. Fair criticism towards the latter includes the already mentioned potential loss of accuracy, especially when compared to standard methods such as FEM. Using the network prediction as initial guess for Newton's method is a way to preserve accuracy while at the same time potentially speed up the iterative process. It is also a natural example of when machine learning frameworks can be combined with existing software for standard methods, thus motivating the learning of the standard method's approximation, here the final element solution. The finite element software we use here is the popular finite element library FEniCS\footnote{\label{fn_hwswdiff} The examples in Section~\ref{sec_nonlinelast} were implemented by a different author than the ones in Sections~\ref{ex:parameter} and \ref{ex:grf}, hence the difference in hardware and software.}~\cite{LoggEtal2012}.

\begin{figure}
\centering
\def\svgwidth{0.8\textwidth}
\begingroup%
  \makeatletter%
  \providecommand\color[2][]{%
    \errmessage{(Inkscape) Color is used for the text in Inkscape, but the package 'color.sty' is not loaded}%
    \renewcommand\color[2][]{}%
  }%
  \providecommand\transparent[1]{%
    \errmessage{(Inkscape) Transparency is used (non-zero) for the text in Inkscape, but the package 'transparent.sty' is not loaded}%
    \renewcommand\transparent[1]{}%
  }%
  \providecommand\rotatebox[2]{#2}%
  \newcommand*\fsize{\dimexpr\f@size pt\relax}%
  \newcommand*\lineheight[1]{\fontsize{\fsize}{#1\fsize}\selectfont}%
  \ifx\svgwidth\undefined%
    \setlength{\unitlength}{484.4491457bp}%
    \ifx\svgscale\undefined%
      \relax%
    \else%
      \setlength{\unitlength}{\unitlength * \real{\svgscale}}%
    \fi%
  \else%
    \setlength{\unitlength}{\svgwidth}%
  \fi%
  \global\let\svgwidth\undefined%
  \global\let\svgscale\undefined%
  \makeatother%
  \begin{picture}(1,0.30451124)%
    \lineheight{1}%
    \setlength\tabcolsep{0pt}%
    \put(0,0){\includegraphics[width=\unitlength,page=1]{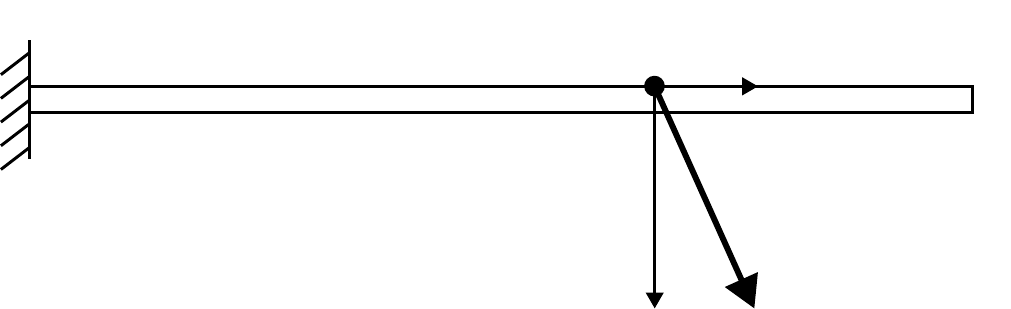}}%
    \put(0.77116393,0.00955637){\makebox(0,0)[lt]{\lineheight{1.25}\smash{\begin{tabular}[t]{l}$F$\end{tabular}}}}%
    \put(0.7402009,0.24177889){\makebox(0,0)[lt]{\lineheight{1.25}\smash{\begin{tabular}[t]{l}$F_x$\end{tabular}}}}%
    \put(0.58538583,0.00955637){\makebox(0,0)[lt]{\lineheight{1.25}\smash{\begin{tabular}[t]{l}$F_y$\end{tabular}}}}%
    \put(0.58538587,0.16437153){\makebox(0,0)[lt]{\lineheight{1.25}\smash{\begin{tabular}[t]{l}$p$\end{tabular}}}}%
    \put(0,0){\includegraphics[width=\unitlength,page=2]{nonlinelast_beamproblem.pdf}}%
    \put(0.58538583,0.27274192){\makebox(0,0)[lt]{\lineheight{1.25}\smash{\begin{tabular}[t]{l}$\text{supp}(F)_p$\end{tabular}}}}%
    \put(0,0){\includegraphics[width=\unitlength,page=3]{nonlinelast_beamproblem.pdf}}%
  \end{picture}%
\endgroup%

\caption{The 2D nonlinear neo-Hookean cantilever beam problem considered in Section~\ref{sec_nonlinelast}. The external force field $F = (F_x, F_y)$ is applied to the surface interval $\text{supp}(F)_p$ with midpoint $p$. The problem parameter tuple is $(p, F_x, F_y)$. The beam and the surface interval are to scale.}
\label{fig_nonlinelast_beamproblem}
\end{figure}

Employing neural networks to predict hyperelasticity solutions has been done in, e.g., \cite{nguyen_deep_2020} with a discretization-free method and in \cite{brink_neural_2021} with a meshfree collocation method. Here we consider a nonlinear neo-Hookean elasticity problem: a two-dimensional cantilever beam under the influence of a given external force. The external force is defined by three variables: the midpoint $p$ of a fixed-length surface interval of application on top of the beam, and the horizontal and vertical magnitudes $F_x$ and $F_y$, respectively. The solution to the problem is the displacement field over the beam. See Figure~\ref{fig_nonlinelast_beamproblem} for an illustration of the problem setup. For the neo-Hookean elasticity model we consider a compressible material with Young's modulus $10^3$ and Poisson's ratio 0.45. We also use a quadratic approximation of the natural logarithm term, i.e., $-2 \ln(J) \approx J^2 - 4J + 3$, where $J = \det(\mathbf{F})$ and $\mathbf{F}$ is the deformation gradient. The beam length and thickness are 0.8 and 0.025, respectively. The application interval of the force is of length 0.084.
We generate a structured uniform triangular 40x2 mesh of the resting beam and consider standard P1 finite elements. This results in 240 DOFs, since every mesh node ($41\cdot 3 = 123$) has two DOFs (one for each dimension) and the DOFs of the left-most three nodes are removed since that is where the beam is fixed.

The solution operator we aim to learn is thus the map from the 3 force variables to the 240 finite element DOFs giving the corresponding displacement of the beam. We use a simple MLP architecture that takes 3 input variables (the force variables), has 4 hidden layers of width 256 with ELU activation and one output layer of width 240 (number of DOFs) with no activation. Just as in the previous examples we use the energy functional corresponding to the PDE as the foundation for the loss function. Here, the energy is
\begin{align}
E(v) = \int_\Omega \Psi(v) \dx - \int_{\partial \Omega \cap \text{supp}(F)_p} F \cdot v \dss
\end{align}
where $v$ is a two-dimensional vector field over the unbent beam (the displacement field), $\Omega \subset \mathbb{R}^2$ is the spatial domain occupied by the unbent beam, $\Psi$ is the neo-Hookean elastic stored energy density, $\partial \Omega$ is the boundary of $\Omega$, $F = (F_x, F_y)$ is the external force field, and $\text{supp}(F)_p$ is the surface interval of application of the force field (the support of $F$) with midpoint $p$. Gravity is not included.

\begin{rem}
An interesting thing we noted when experimenting with this setup was the limiting effect of the considered input. For instance, we tried to train networks that mapped 4 force variables (2 different forces, each with a $p$ and an $F_y$) to a corresponding displacement field. For this case, one can place a downward force at the free end of the beam and an upward force in the middle, which should result in a beam that is bent like a hill. However, even though we tested with various hyperparameters and settings for the network and training, our method failed to capture the desired physics to a satisfactory degree. For comparison, in \cite{BURMAN2025118111} where we considered a nonlinear Poisson energy, we managed to use 21 input parameters and still capture the physics. We suspect that this difference and difficulty could be related to the increased nonlinear nature of neo-Hookean elasticity compared to the nonlinear Poisson energy. The nonlinearity in the latter comes from the coefficient $A(v) = 1 + v^2$ which still may be considered fairly mild since it makes the problem resemble the corresponding linear one and behave well (See the assumptions in Lemma~\ref{lem:nonlinenergyident} and Example~\ref{ex_nonlinen_r}). A final remark is that, in our experience, the size of the input has a much greater effect on the solution operator network's approximation capabilities than the output size (This relates to the below discussion about the results in Table~\ref{table_nonlinelast_learnerrs_eb}).
\end{rem}

Back to the problem at hand. During training, we pick 32 randomly selected tuples of force variables, i.e., $32 \times (p, F_x, F_y)$, in each iteration to compute an average loss. The midpoint of the interval of application is sampled uniformly along the beam $p \sim \mcU = \mcU([x_{\min}, x_{\max}])$, where $x_{\min}$ and $x_{\max}$  are the left and right endpoints of the beam, respectively. The horizontal and vertical magnitudes are sampled from normal distributions with mean 0 and standard deviation $\sigma$, i.e., $F_x, F_y \sim \mcN(0, \sigma^2)$. The optimization is performed with the Adam optimizer where we perform $10^6$ iterations with a decreasing learning rate. The learning rate starts at $10^{-4}$ and after every 250k iterations it is decreased by a factor of 0.5.

Sampling the force magnitudes from normal distributions with mean 0 means that we train the networks more for problems with solutions close to the zero function, i.e., the unbent beam. It is thus reasonable to expect the networks to perform better for such problems than for problems with large forces resulting in large deformations. The previously used relative learning errors are thus unsuitable here, since they could potentially give very large errors for problems with solutions close to zero and smaller errors for more deformed solutions. However, some kind of error normalization is typically desired to make it easier to interpret the numerical values. The previously used relative learning errors are normalized with respect to both the reference solution $u_h$ and the domain $\Omega$ (via both the energy and the norms). As explained above, reference solution normalization is unsuitable here so therefore we only normalize with respect to $\Omega$. Letting $\bfF = (p, F_x, F_y)$ and $|\Omega|$ denote the Lebesgue measure of $\Omega$, we thus compute the \emph{$\Omega$-relative} learning errors
\begin{align}
	&|E(\mcA_{h,\theta}(\bfF_i)) - E(\mcA_{h}(\bfF_i))|/|\Omega| \\
	&\|\mcA_{h,\theta}(\bfF_i) - \mcA_{h}(\bfF_i)\|_{L^2}/|\Omega|^{1/2} \\
	&\|\mcA_{h,\theta}(\bfF_i) - \mcA_{h}(\bfF_i)\|_{H_0^1}/|\Omega|^{1/2}
\end{align}

We consider two different force situations: basic beam bending and extreme bending.

\subsubsection{Basic bending}

Here, we only consider one simple sampling case for the force variables in which we do not include a horizontal magnitude. The force variables are sampled by
\begin{equation}
p \sim \mcU \quad F_x = 0 \quad F_y \sim \mcN(0, 1)
\end{equation}
Training and inference times for the basic bending case on both a Nvidia A100 GPU and an Apple M1 CPU\footnote{\label{fn_hwswdiff_2} See footnote \ref{fn_hwswdiff}.} are presented in Table~\ref{table_nonlinelast_traintimes_bb}. The training times clearly show the benefit of GPU training.

\begin{table}[h!]
	\caption{Training and inference times for the basic bending case.}
	\centering
	\begin{tabular}{|l|l|l|l|l|}
		\hline
		 Hardware & Training time & Inference time \\ \hline
		 GPU (A100) & 558 s & 0.8 ms \\ \hline
		 CPU (M1) & 2393 s & 0.13 ms \\ \hline
	\end{tabular}\label{table_nonlinelast_traintimes_bb}
\end{table}

\begin{figure}[h!]
	\centering
	\begin{subfigure}[b]{0.35\textwidth}
		\includegraphics[width=\textwidth]{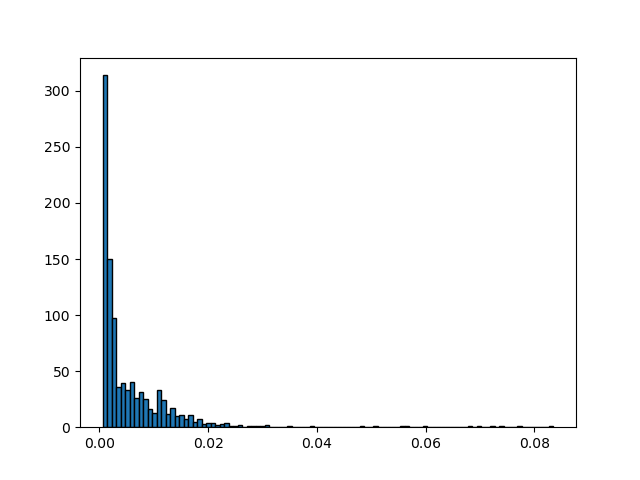}
		\caption{Energy}
	\end{subfigure}\hspace{-1.4em}
	\begin{subfigure}[b]{0.35\textwidth}
		\includegraphics[width=\textwidth]{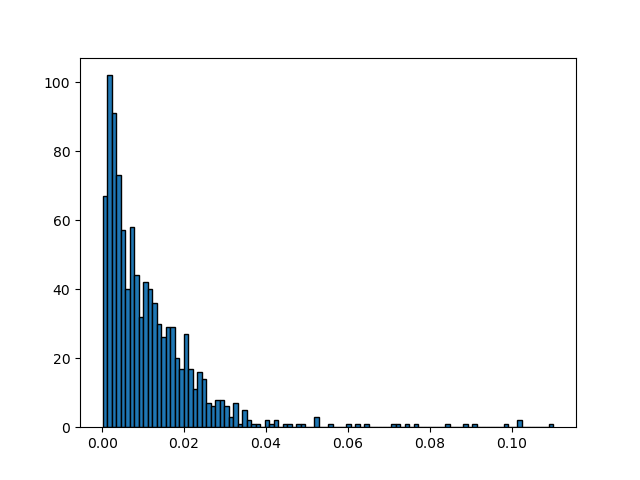}
		\caption{$L^2$-norm}
	\end{subfigure}\hspace{-1.4em}
	\begin{subfigure}[b]{0.35\textwidth}
		\includegraphics[width=\textwidth]{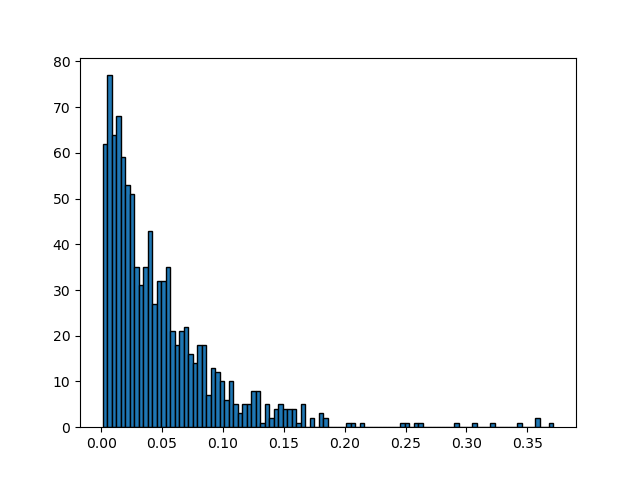}
		\caption{$H_0^1$-norm}
	\end{subfigure}
	\caption{Histogram plots for the different $\Omega$-relative learning errors for the basic bending case, i.e., Table~\ref{table_nonlinelast_learnerrs_bb}. Sample size $K = 10^3$.} \label{fig_nonlinelast_bb_histograms}
\end{figure}

Learning errors for the basic bending case are presented in Table~\ref{table_nonlinelast_learnerrs_bb} and corresponding histogram plots are presented in Figure~\ref{fig_nonlinelast_bb_histograms}. The histograms show that the error distributions are skewed towards zero, indicating some degree of well-trainedness.

\begin{table}[h!]
	\caption{$\Omega$-relative learning errors for the basic bending case. Sample size $K = 10^3$.}
	\centering
	{
	\sisetup{scientific-notation=true, output-exponent-marker = e, table-format=1.1e1, table-text-alignment=left, round-mode=places, round-precision=1}
	\begin{tabular}{| l | l | S | S | S |}
		\hline
		Training \& test data & Measure & {Energy} & {$L^2$-norm} & {$H_0^1$-norm}  \\ \hline
		\multirow{2}{*}{$p \sim \mcU \quad F_x = 0 \quad F_y \sim \mcN(0, 1)$} & Mean & 0.0058 & 0.0115 & 0.0477 \\ \cline{2-5}
											& Std & 0.0085 & 0.0126 & 0.0481 \\ \hline
	\end{tabular}\label{table_nonlinelast_learnerrs_bb}
	}
\end{table}

\begin{figure}[h!]
	\centering
	\begin{subfigure}[b]{0.49\textwidth}
		\includegraphics[width=\textwidth]{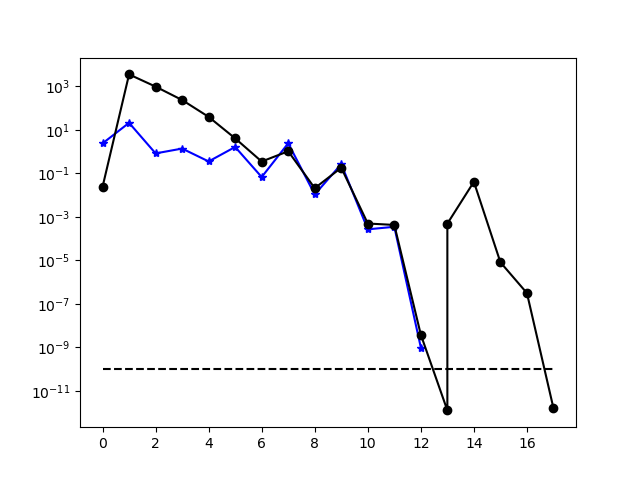}
		\caption{Absolute residuals ($r_n$) against iterations ($n$). Tolerance $= 10^{-10}$.}
	\end{subfigure}~
	\begin{subfigure}[b]{0.49\textwidth}
		\includegraphics[width=\textwidth]{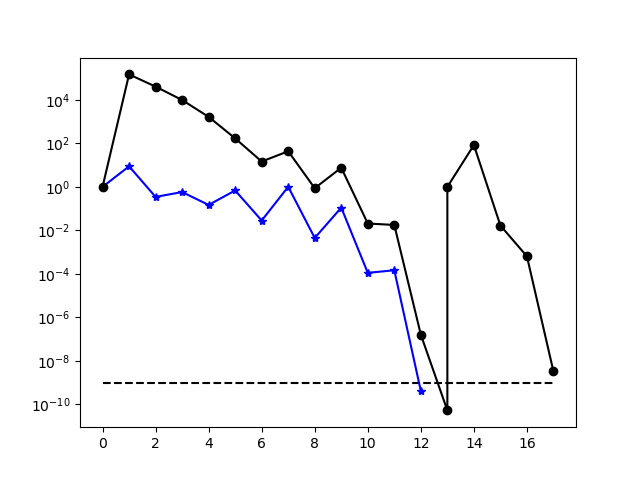}
		\caption{Relative residuals ($r_n/r_0$) against iterations ($n$). Tolerance $= 10^{-9}$.}
	\end{subfigure}
	\caption{Newton iteration convergence plots for a \emph{basic bending} case. Newton residuals (absolute in (a), relative in (b)) for using the network prediction (blue, stars) and the zero function (black, dots) as initial guesses are plotted against iterations. The dashed lines are the default tolerances used by FEniCS. Whichever of the absolute and relative tolerance is reached first determines convergence. Using the network prediction results in 12 iterations and the zero function in a total of 17. When using the zero function, the iterations had to be split up in a two-part load stepping to converge. This is the reason for the sharp rise in the black dot curves at iteration 13. The break point for this load stepping was chosen optimally to reduce the total iteration time as much as possible. Trial and error gave the break point to be 98\% of the full force magnitude.}
	\label{fig_nonlinelast_bb_newtonconv}
\end{figure}

\begin{figure}[h!]
	\centering
	\begin{subfigure}[b]{0.49\textwidth}
		\includegraphics[width=\textwidth]{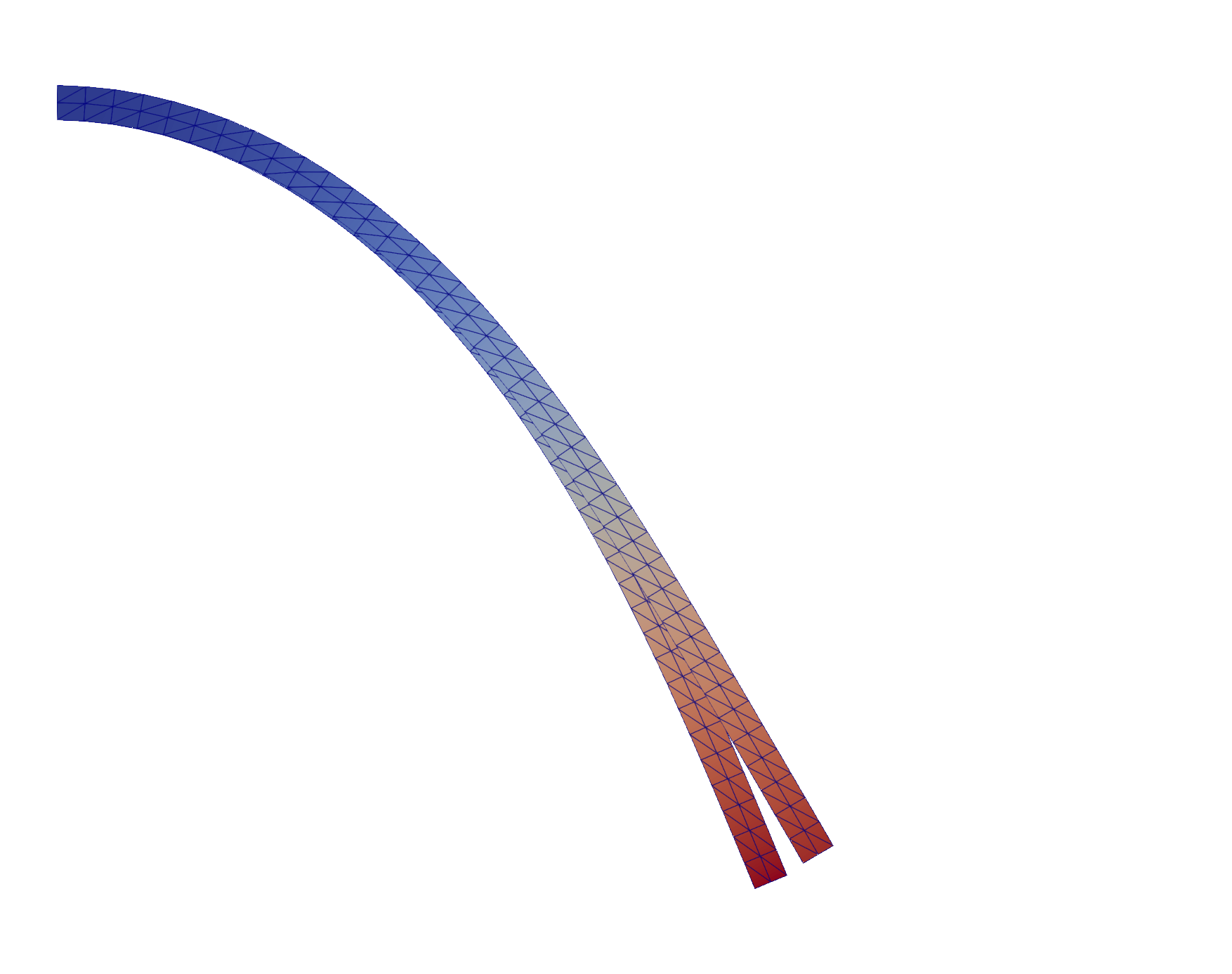}
		\caption{Network prediction (the upper beam) as initial guess.}
	\end{subfigure}~
	\begin{subfigure}[b]{0.49\textwidth}
		\includegraphics[width=\textwidth]{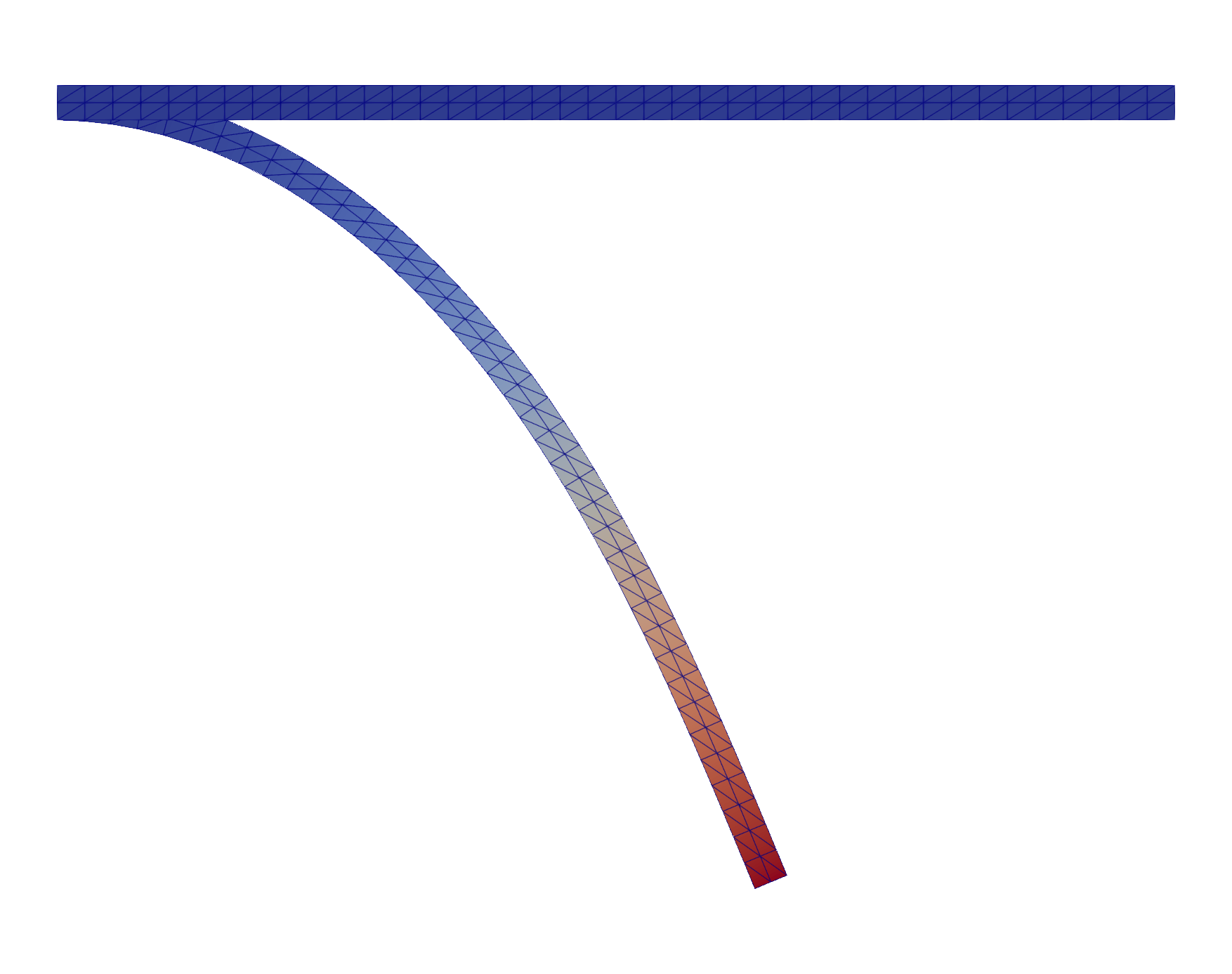}
		\caption{Zero function (the unbent beam) as initial guess.}
	\end{subfigure}
	\caption{Different initial guesses for Newton's method with resulting converged solutions for a \emph{basic bending} case. The converged solution $u_h$ can be considered the same (within given tolerances) in both cases (the lower beam in both (a) and (b)).}
	\label{fig_nonlinelast_bb}
\end{figure}

We now consider the specific problem of computing the displacement when the external force is located at the right end of the beam and directed downwards with magnitude 1.0. We use the operator network for the basic bending case and perform the Newton iterations with a FEniCS implementation of the problem. All computations are performed on an Apple M1 CPU. Results from using the network prediction and a standard initial guess (the zero function) as input for Newton's method are presented in Table~\ref{table_nonlinelast_newtontimes_bb} and Figures~\ref{fig_nonlinelast_bb_newtonconv} and \ref{fig_nonlinelast_bb}.

\begin{table}[h!]
	\caption{Computational times on an Apple M1 CPU for using different initial guesses for Newton's method.}
	\centering
	\begin{tabular}{|l|l|l|l|l|}
		\hline
		Initial guess & $E(u_h)$ & Inference time & Newton time & Total time  \\ \hline
		Network prediction & -0.014397 & 0.094 ms & 15.678 ms & 15.772 ms \\ \hline
		Zero function & -0.014397 & -- & 22.658 ms & 22.658 ms \\ \hline
	\end{tabular}\label{table_nonlinelast_newtontimes_bb}
\end{table}

The results in Table~\ref{table_nonlinelast_newtontimes_bb} and Figures~\ref{fig_nonlinelast_bb_newtonconv} and \ref{fig_nonlinelast_bb} show that using the network prediction as initial guess for Newton's method preserves accuracy while at the same time provides a speed-up compared to using the standard zero function as initial guess. However, to properly benefit from using the network prediction as initial guess, a large enough number of problems will have to be solved so that all individual problem gains compensate for the training time. From Tables~\ref{table_nonlinelast_traintimes_bb} and \ref{table_nonlinelast_newtontimes_bb}, we compute that one needs to solve at least 81,034 problems with GPU training and 347,517 with CPU training to get this benefit. For comparison, if the network prediction is already considered good enough, i.e., only performing inference without subsequent Newton iterations, then the corresponding numbers are 24,730 problems with GPU training and 106,054 with CPU training. This might sound like a lot, but if the trained network is incorporated into software intended for many users or if many instances have to be solved, e.g., in an optimization process or for a time-dependent version of the problem, the numbers quickly diminish in perspective.

\subsubsection{Extreme bending} \label{sec_extremebending}

Here, we consider three more advanced sampling cases for the force variables:
\begin{align}
& p \sim \mcU \quad F_x, F_y \sim \mcN(0, 1) \\
& p \sim \mcU \quad F_x, F_y \sim \mcN(0, 4) \\
& p \sim \mcU \quad F_x, F_y \sim \mcN(0, 16) \label{eq_traindata_eb_case3}
\end{align}
Training and inference times for the extreme bending cases on both an A100 GPU and an Apple M1 CPU are presented in Table~\ref{table_nonlinelast_traintimes_eb}. Again the results show the benefits of GPU training. We note that the training times for the three CPU cases vary much more than the corresponding GPU times. We do not think that the different sampling cases are the reason for this. Instead, the CPU training times should probably be considered belonging to the same range of possible values, i.e., the variation should not be given any greater significance. This also applies to the CPU inference times and the GPU times.

\begin{table}[h!]
	\caption{Training and inference times for the extreme bending cases. The A100 GPU Utilization was 46\% in all three cases.}
	\centering
	\begin{tabular}{|l|l|l|l|l|}
		\hline
		 Training \& test data & Hardware & Training time & Inference time \\ \hline
		 \multirow{2}{*}{$p \sim \mcU \quad F_x, F_y \sim \mcN(0, 1)$} & GPU (A100) & 583 s & 0.85 ms \\ \cline{2-4}
		 								       & CPU (M1) & 2398 s & 0.17 ms \\ \hline
		 \multirow{2}{*}{$p \sim \mcU \quad F_x, F_y \sim \mcN(0, 4)$} & GPU (A100) & 583 s & 0.88 ms \\ \cline{2-4}
		 								       & CPU (M1) & 2438 s & 0.14 ms \\ \hline
		 \multirow{2}{*}{$p \sim \mcU \quad F_x, F_y \sim \mcN(0, 16)$} & GPU (A100) & 582 s & 0.86 ms \\ \cline{2-4}
		 								       & CPU (M1) & 2672 s & 0.13 ms \\ \hline
	\end{tabular}\label{table_nonlinelast_traintimes_eb}
\end{table}

\begin{figure}[h!]
	\centering
	\begin{subfigure}[b]{0.35\textwidth}
		\includegraphics[width=\textwidth]{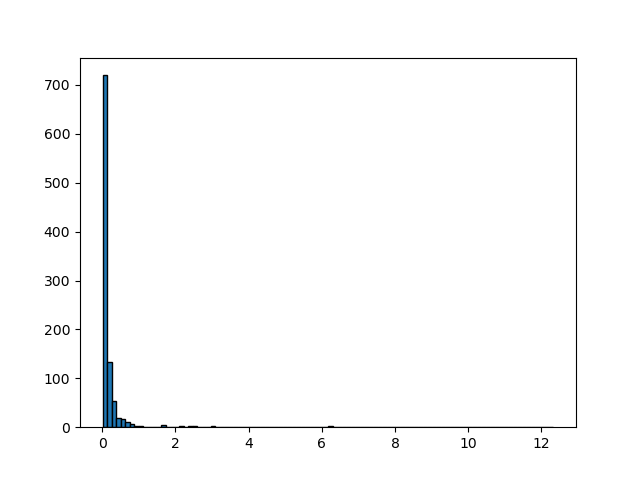}
		\caption{Energy}
	\end{subfigure}\hspace{-1.4em}
	\begin{subfigure}[b]{0.35\textwidth}
		\includegraphics[width=\textwidth]{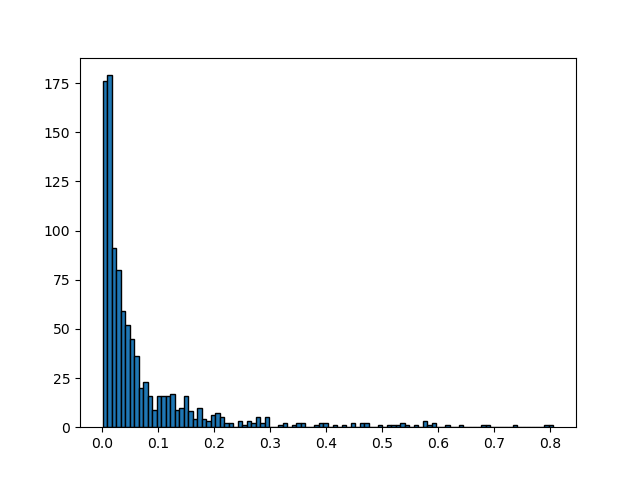}
		\caption{$L^2$-norm}
	\end{subfigure}\hspace{-1.4em}
	\begin{subfigure}[b]{0.35\textwidth}
		\includegraphics[width=\textwidth]{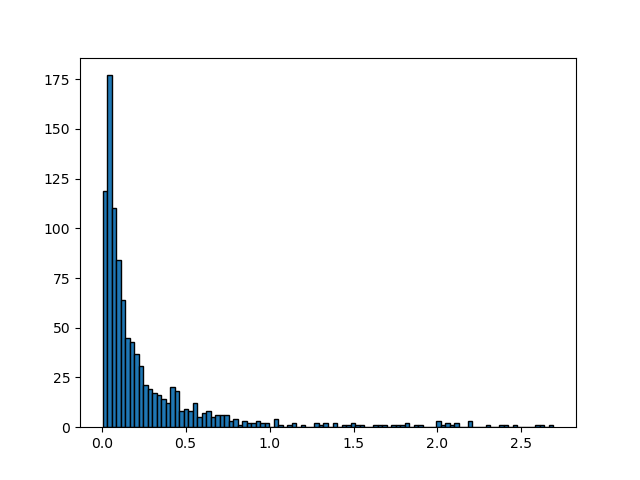}
		\caption{$H_0^1$-norm}
	\end{subfigure}
	\caption{Histogram plots for the different $\Omega$-relative learning errors for the last extreme bending case, i.e., the last row of Table~\ref{table_nonlinelast_learnerrs_eb}. Sample size $K = 10^3$.} \label{fig_nonlinelast_eb_histograms}
\end{figure}

Learning errors for the extreme bending cases are presented in Table~\ref{table_nonlinelast_learnerrs_eb} and histograms for the last case are presented in Figure~\ref{fig_nonlinelast_eb_histograms}. The learning errors increase with increased variance of the sampling distributions. This is reasonable since wider distributions can be thought of as larger input domains of the map that the network approximates. Larger input domains mean that the network is expected to be able to handle more cases. This in turn means that the network could be harder to train well for the given task. Nevertheless, the histograms for the last case all show small error distributions skewed towards zero, indicating some degree of well-trainedness.

\begin{table}[h!]
	\caption{$\Omega$-relative learning errors for the extreme bending cases. Sample size $K = 10^3$.}
	\centering
	{
	\sisetup{scientific-notation=true, output-exponent-marker = e, table-format=1.1e1, table-text-alignment=left, round-mode=places, round-precision=1}
	\begin{tabular}{| l | l | S | S | S |}
		\hline
		Training \& test data & Measure & {Energy} & {$L^2$-norm} & {$H_0^1$-norm}  \\ \hline
		\multirow{2}{*}{$p \sim \mcU \quad F_x, F_y \sim \mcN(0, 1)$} & Mean & 0.0443 & 0.0395 & 0.1564 \\ \cline{2-5}
										& Std & 0.1720 & 0.0749 & 0.2928 \\ \hline
		\multirow{2}{*}{$p \sim \mcU \quad F_x, F_y \sim \mcN(0, 4)$} & Mean & 0.1396 & 0.0574 & 0.2236 \\ \cline{2-5}
										& Std & 0.4489 & 0.1000 & 0.3819 \\ \hline
		\multirow{2}{*}{$p \sim \mcU \quad F_x, F_y \sim \mcN(0, 16)$} & Mean & 0.2299 & 0.0704 & 0.2669 \\ \cline{2-5}
										& Std & 0.7610 & 0.1104 & 0.4052 \\ \hline
	\end{tabular}\label{table_nonlinelast_learnerrs_eb}
	}
\end{table}

\begin{figure}[h!]
	\centering
	\begin{subfigure}[b]{0.49\textwidth}
		\includegraphics[width=\textwidth]{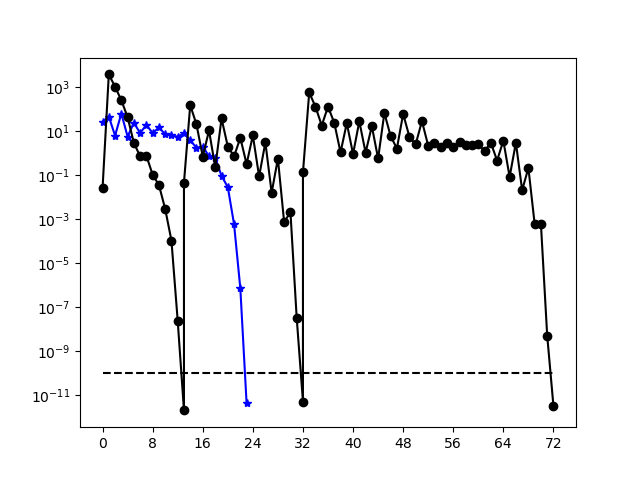}
		\caption{Absolute residuals ($r_n$) against iterations ($n$). Tolerance $= 10^{-10}$.}
	\end{subfigure}~
	\begin{subfigure}[b]{0.49\textwidth}
		\includegraphics[width=\textwidth]{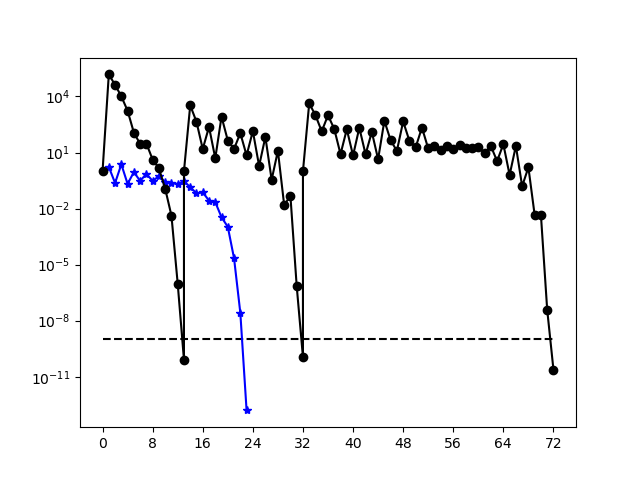}
		\caption{Relative residuals ($r_n/r_0$) against iterations ($n$). Tolerance $= 10^{-9}$.}
	\end{subfigure}
	\caption{Newton iteration convergence plots for an \emph{extreme bending} load stepping case. Newton residuals (absolute in (a), relative in (b)) for using the network intermediate (blue, stars) and the zero function (black, dots) as initial guesses are plotted against iterations. The dashed lines are the default tolerances used by FEniCS. Whichever of the absolute and relative tolerance is reached first determines convergence. Using the network intermediate results in 23 iterations and the zero function in a total of 72. When using the zero function, the iterations are split up in a three-part load stepping. This is the reason for the sharp rises in the black dot curves at iteration 13 and 32.}
	\label{fig_nonlinelast_eb_newtonconv}
\end{figure}

\begin{figure}[h!]
	\centering
	\begin{subfigure}[b]{0.49\textwidth}
		\includegraphics[width=\textwidth]{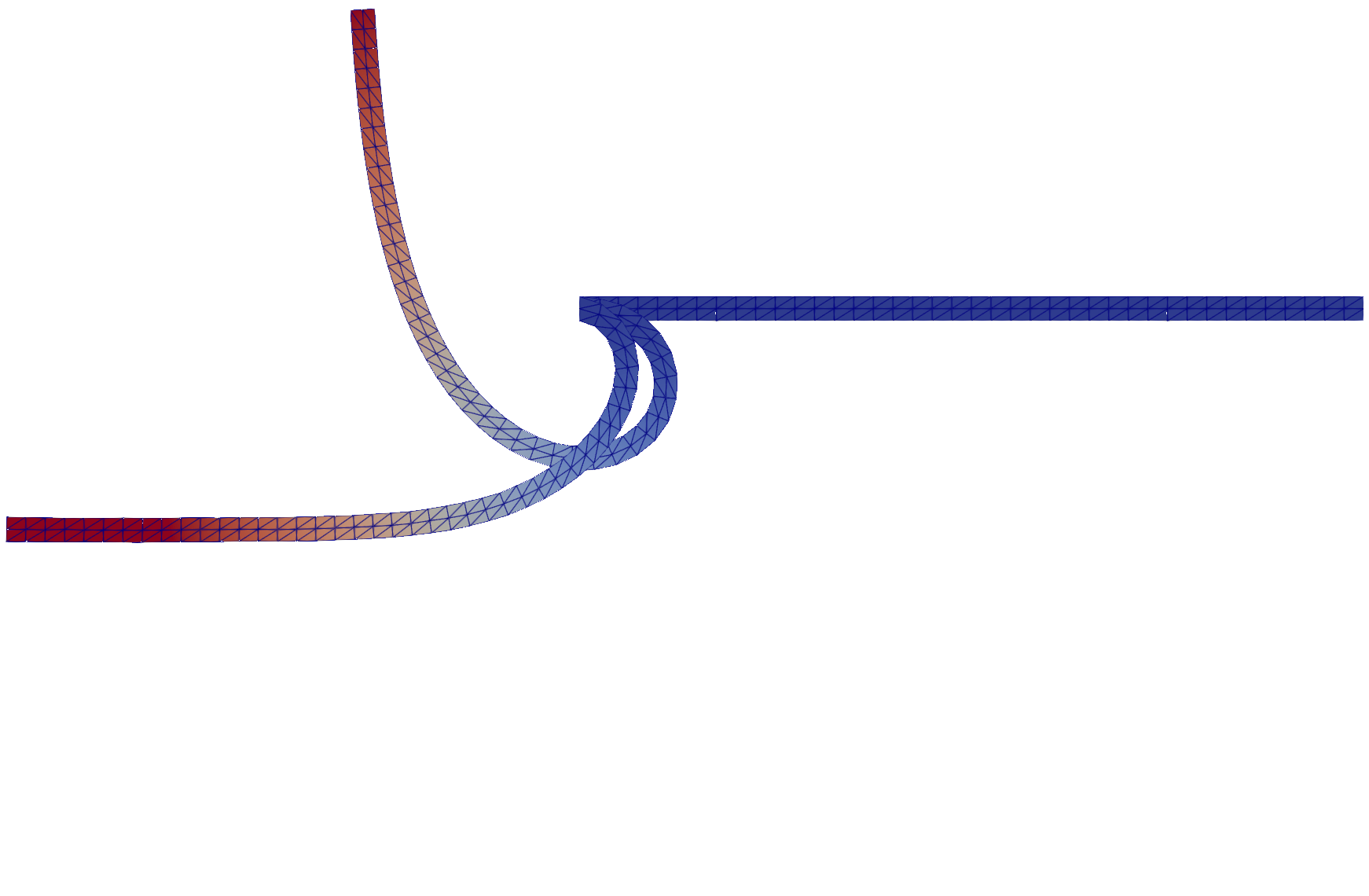}
		\caption{Network intermediate (the leftward pointing beam) as initial guess. The unbent beam is present for reference.}
	\end{subfigure}~
	\begin{subfigure}[b]{0.49\textwidth}
		\includegraphics[width=\textwidth]{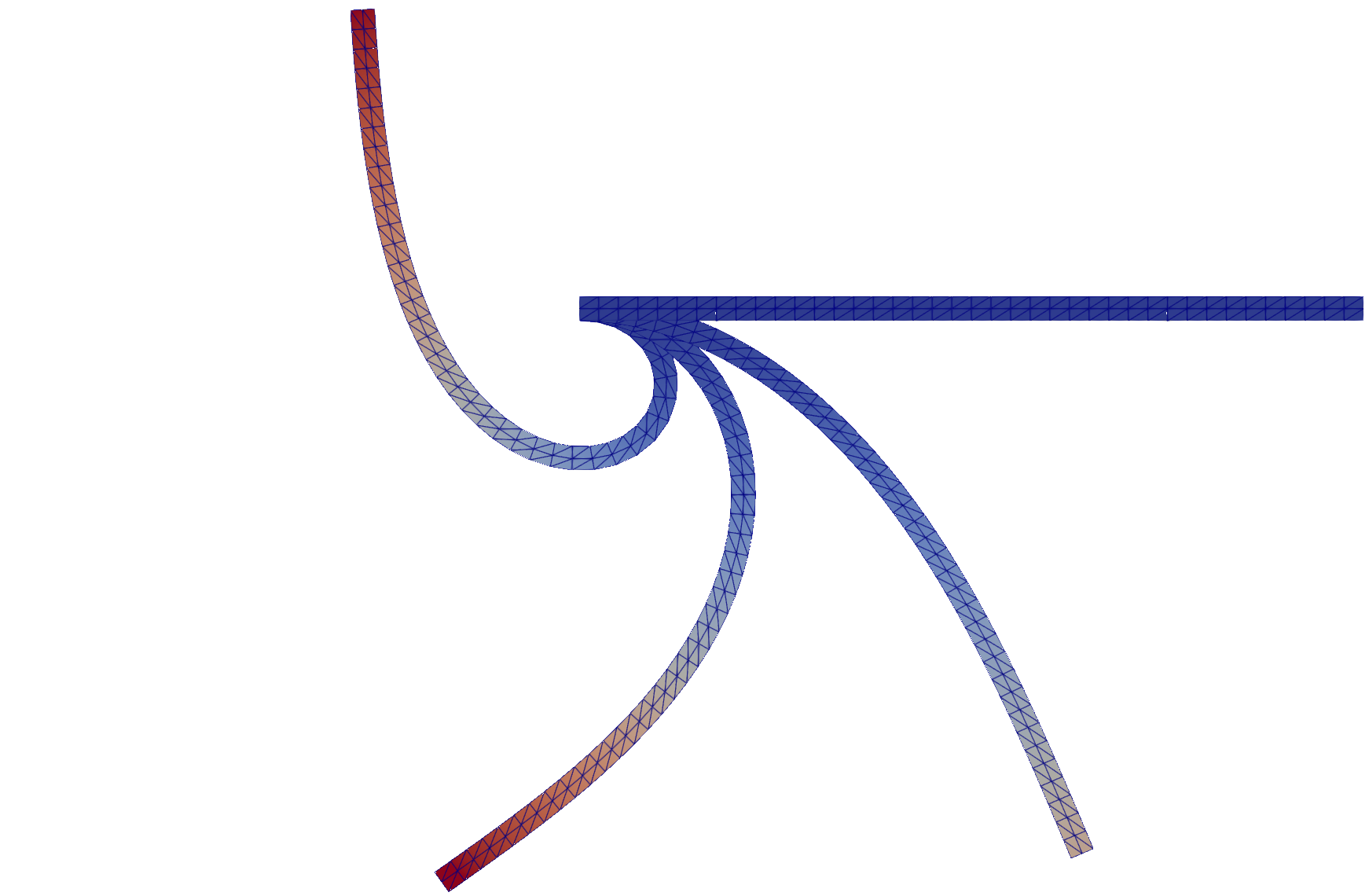}
		\caption{Zero function (the unbent beam) as initial guess. The two intermediate states of the load stepping are also present.}
	\end{subfigure}
	\caption{Different initial guesses for Newton's method with resulting converged solutions for an \emph{extreme bending} load stepping case. The converged solution $u_h$ can be considered the same (within given tolerances) in both cases (the upward pointing beam in both (a) and (b)). The network intermediate state comes from an operator network where $F_x, F_y \sim \mcN(0, 16)$ during training.}
	\label{fig_nonlinelast_eb}
\end{figure}

We now consider a load stepping problem. The specific problem is to compute the displacement when the external force is again located at the right end of the beam but successively turns around the fixed end, forcing the beam to curl. This is different from the basic bending example since the force now \emph{changes} when the beam has been bent to a certain degree. Doing this with the standard approach means performing \emph{sets} of Newton iterations: From an initial state of the beam and an initial force, Newton iterations are performed until convergence. The converged state is then used as the new starting state together with a new force, and so on. The last converged state is the desired one and the other ones are simply intermediates. The forces we consider for the standard approach are:
\begin{align}
F_1 = (0.0, -0.99) \quad F_2 = (-1.5, 0.0) \quad F_3 = (0.0, 5.0)
\end{align}
This means that we will get three states: two intermediates and one final curled state. For the machine learning approach, we note that the operator networks we have trained cannot reach the final curled state of the beam. This is because the forces used as input to the network are all applied to the unbent beam. However, the operator networks can still provide predictions of intermediate states that can be useful as initial guesses. We use an operator network from the last extreme bending case, i.e., where $F_x, F_y \sim \mcN(0, 16)$ during training, and take the network intermediate prediction coming from $F = (-10.0, -5.0)$ as initial guess. We again perform the Newton iterations with a FEniCS implementation of the problem. All computations are performed on an Apple M1 CPU. Results from using the network intermediate state and the successive beam configurations starting from the standard initial guess (the zero function) as input for Newton's method are presented in Table~\ref{table_nonlinelast_newtontimes_eb} and Figures~\ref{fig_nonlinelast_eb_newtonconv} and \ref{fig_nonlinelast_eb}.

\begin{table}[h!]
	\caption{Computational times on an Apple M1 CPU for using different initial guesses for Newton's method. The network intermediate state comes from an operator network where $F_x, F_y \sim \mcN(0, 16)$ during training.}
	\centering
	\begin{tabular}{|l|l|l|l|l|}
		\hline
		Initial guess & $E(u_h)$ & Inference time & Newton time & Total time  \\ \hline
		Network intermediate & 0.067984 & 0.095 ms & 19.925 ms & 20.02 ms \\ \hline
		Zero function & 0.067984 & -- & 47.84 ms & 47.84 ms \\ \hline
	\end{tabular}\label{table_nonlinelast_newtontimes_eb}
\end{table}

The results in Table~\ref{table_nonlinelast_newtontimes_eb} and Figures~\ref{fig_nonlinelast_eb_newtonconv} and \ref{fig_nonlinelast_eb} show that just as in the basic bending case, using the network output as initial guess for Newton's method preserves accuracy while at the same time provides a speed-up compared to using the standard zero function as initial guess. Furthermore, the gain from using the network output is higher in the extreme bending case compared to the basic one. This is easily seen by comparing the total times in Tables~\ref{table_nonlinelast_newtontimes_bb} and \ref{table_nonlinelast_newtontimes_eb}. For the basic bending case we have that the total time with the network output is about 70\% of the total time with the zero function. The corresponding quantity for the extreme bending case is 42\%. This higher gain naturally affects the number of problems that have to be solved in order to compensate for the training time of the network. From the last case in Table~\ref{table_nonlinelast_traintimes_eb} and the times in Table~\ref{table_nonlinelast_newtontimes_eb}, we compute that one needs to solve at least 20,921 problems with GPU training and 96,047 with CPU training to get a benefit from using the network output as initial guess when taking the training time into account. These numbers for the extreme bending case are 26\% and 28\% of those of the basic bending case for GPU and CPU training, respectively. We point out that the higher gain from using a network in an extreme bending case could in fact be even higher. This is so since here we only use the network to get an \emph{intermediate} state, not a prediction of the solution as in the basic bending case. From Tables~\ref{table_nonlinelast_newtontimes_bb} and \ref{table_nonlinelast_newtontimes_eb}, the Newton solver time for using a prediction in the basic bending case is 79\% of the one for using an intermediate state in the extreme case, suggesting the even higher computational gain. As a final remark we point out that the two Newton solver times do not seem to differ that much while the corresponding beam configuration pairs (starting and converged states) differ quite a lot visually: From Figures~\ref{fig_nonlinelast_bb} and \ref{fig_nonlinelast_eb}, the starting state is much closer to the converged one in the basic bending case compared to in the extreme one. We take this to mean that most of the Newton computational work is performed towards the end, fine-tuning the beam configuration to the final state.

To summarize, combining operator networks with Newton's method preserves accuracy and can provide a speed-up. In the case of nonlinear elasticity, the greater the deformation, the greater the potential speed-up from using operator networks.

\section{Conclusions and outlook}

We have presented a machine learning framework for learning solutions to a class of PDE problems. A core idea of the framework is to learn the corresponding discrete solution of some standard numerical method instead of aiming for the exact solution. The reason being that the standard method could be used to aid and enhance the framework. This core idea can in general be applied to various machine learning frameworks and standard methods but here we have considered a simple MLP-architecture together with energy minimization for the framework and FEM as the standard numerical method. We have presented both theoretical results (approximation error estimate) and practical applications (Newton's method) that demonstrate how the framework may be beneficially combined with FEM. We have also presented pure framework results that show strengths and limitations of it as well as potential applications. These results are the learning errors, the usage of batches of elements for the energy during training, and the computation of quantities of interest.

Two natural avenues for future work are more advanced network architectures and training algorithms. For example, the neural operators in \cite{MR4582511} use lifting and projection maps to transform input and output in order for the same network to handle different domain discretizations. This is a relevant extension for our method since in its current form, the mesh used is hardcoded into the network via the output layer. Concerning training, a possible improvement could be to use so-called energy natural gradient descent \cite{muller23b} which has been able to reduce the learning error by several orders of magnitude compared to, e.g., Adam when training PINNs. In the training we also propose to use random batches of mesh elements when computing the loss function. This batching technique could be further investigated by considering smaller subdomains of elements or by exploring memory access patterns of the elements. The last elasticity example, the extreme bending case presented in Section~\ref{sec_extremebending}, also provides a starting point for future work. We note that in this example, the neural network is limited by the fact that the external forces are always applied to the initial state (the unbent beam). This could be improved by introducing time dependency where the forces are allowed to change during the bending process. This would also most likely mean that the network needs to take the current position of the beam as an input parameter. This is an interesting path for future research.

\bigskip
\paragraph{Acknowledgement} This research was supported in part by the Swedish Research Council
Grants Nos.\ 2021-04925, 2022-03543, 2025-05562, Knut and Alice Wallenberg Foundation,
Grant No.\ KAW 2025.0277, and the Swedish Research Program Essence.

The GPU computations were enabled by resources provided by the National Academic Infrastructure for Supercomputing in Sweden (NAISS), partially funded by the Swedish Research Council through grant agreement No. 2022-06725.

\bibliographystyle{abbrv}
\footnotesize{
\bibliography{stoc-min}
}

\bigskip
\bigskip
\noindent
\footnotesize {\bf Authors' addresses:}

\smallskip
\noindent

\smallskip
\noindent
Mats G. Larson,  \quad \hfill \addressumushort\\
{\tt mats.larson@umu.se}

\smallskip
\noindent
Carl Lundholm, \quad \hfill \addressumushort\\
{\tt carl.lundholm@umu.se}

\smallskip
\noindent
Anna Persson, \quad \hfill Information Technology-Scientific Computing,  Uppsala University, Sweden\\
{\tt apersson@it.uu.se}
\end{document}